\documentclass{article}

\usepackage{PRIMEarxiv}

\usepackage{wrapfig}
\usepackage{times}
\usepackage{epsfig}
\usepackage{graphicx}
\usepackage{amsmath}
\usepackage{amssymb}
\usepackage{amsfonts}       
\usepackage[export]{adjustbox}
\usepackage{subcaption}
\usepackage{booktabs}       
\usepackage{multirow,tabularx}
\usepackage{enumitem}
\usepackage[dvipsnames,svgnames]{xcolor}
\usepackage{cite}

\usepackage{caption}
\usepackage{relsize}
\usepackage{xspace}
\usepackage{tikz}
\usepackage{array}

\definecolor{yellow}{RGB}{230, 234, 44}

\newcommand{\myparagraphbase}[1]{\paragraph{#1}}
\newcommand{\myparagraph}[1]{\paragraph{#1.}}

\makeatletter
\DeclareRobustCommand\onedot{\futurelet\@let@token\@onedot}
\def\@onedot{\ifx\@let@token.\else.\null\fi\xspace}
\def\eg{\emph{e.g}\onedot} 
\def\ie{\emph{i.e}\onedot}

\makeatother
\title{Domain Adaptation in Multi-View Embedding \\ for Cross-Modal Video Retrieval}

\vspace{-10pt}
\author{
Jonathan Munro \\ University of Bristol \And %
Michael Wray \\ University of Bristol \And %
Diane Larlus \\ NAVER LABS Europe \AND 
Gabriela Csurka \\ NAVER LABS Europe \And 
Dima Damen \\ University of Bristol \AND 
}

\date{}
\begin{document}
\maketitle

\begin{figure*}[h!]
    \includegraphics[width=1\linewidth]{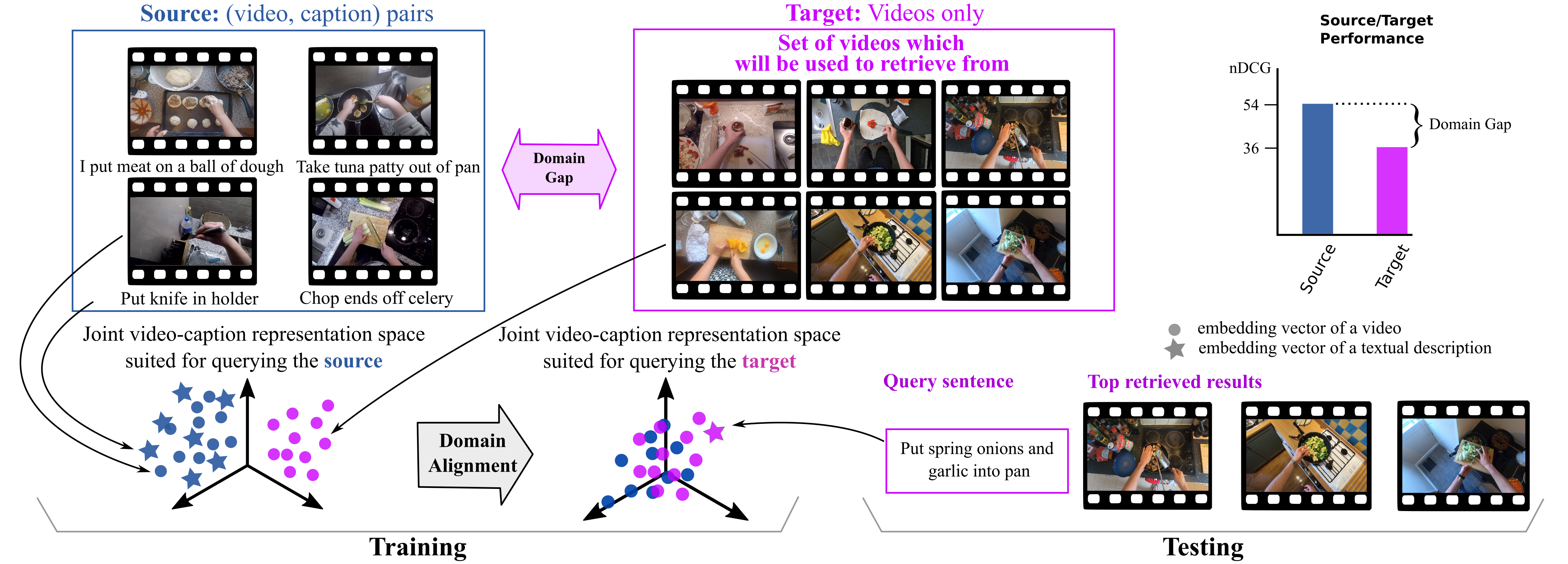}
    {\caption{Given video-text pairs (respectively denoted by circles and stars) from the {\color{blue} source (blue)}, and a video-only {\color{violet}target set (purple)}, we propose an alignment method to reduce the domain gap between the source videos and the target videos using pseudo-labels (Section~\ref{sec:pseudo}) and cross-domain ranking (Section~\ref{sec:cross-domain-embedd}). The learnt and aligned space can then be used for retrieving a ranked list of target videos using previously unseen text queries.}\label{fig:front_page}}
    \vspace{2cm}
\end{figure*}

\begin{abstract}
Given a gallery of uncaptioned video sequences, this paper considers the task of retrieving videos based on their relevance to an unseen text query. 
To compensate for the lack of annotations, we rely instead on a related video gallery composed of video-caption pairs, termed the source gallery, albeit with a domain gap between its videos and those in the target gallery.
We thus introduce the problem of Unsupervised Domain Adaptation for Cross-modal Video Retrieval, along with a new benchmark on fine-grained actions.
We propose a novel iterative domain alignment method by means of pseudo-labelling target videos and cross-domain (\ie source-target) ranking.
Our approach adapts the embedding space to the target gallery, consistently outperforming source-only as well as marginal and conditional alignment methods.
\end{abstract}
 
\section{Introduction}
Using natural language queries to search for video sequences is the most intuitive
interface for video search engines.
Example applications include finding short and precise action sequences in
long instructional videos, \eg for educational purposes. 
Cross-modal retrieval is typically approached by training a joint
video-text embedding following a learning-to-rank objective.
Thanks to this, retrieval
comprises of computing distances between representations of textual queries   and representations of videos in a gallery. 
Building a reliable joint embedding space requires large amounts of accurately
annotated video-caption 
pairs~\cite{chen2020fine,gabeur2020multi,hahn2019action2vec,liu2019use,miech2020end,miech2019howto100m,mithun2018learning,wray2019fine},    
where captions come from manual annotations, query logs or transcribed audio.
These works assume the training set allows for learning a robust representation that generalises to the test/target videos.
However, the appearance of certain actions can change due to new environments or recording equipment.
This introduces a visual domain gap between the captioned video sequences and the gallery of videos we wish to retrieve from, rendering the trained joint embedding
space sub-optimal.

In this paper we address the challenge of a visual
domain gap in the context of cross-modal action
retrieval, assuming that no captions are available for the target gallery.
For this, we choose the
EPIC-KITCHENS-100 dataset~\cite{Damen2020RESCALING} which is 
composed of two distinct sets of videos featuring fine-grained actions. 
It has been shown to exhibit a significant domain shift which alters performance when tested for action recognition~\cite{Munro_2020_CVPR,
planamente2021cross,Song_2021_CVPR}.
Instead, we offer the first attempt to use this dataset for domain alignment in action retrieval.

We first show that the marginal alignment of the distribution over videos~\cite{LongICML15LearningTransferableFeaturesDAN, coralAAAI2016}, independent of the text, is insufficient to
fix the domain gap. 
Instead, we propose to leverage pseudo-labels for the target videos in a unified learning-to-rank approach which effectively combines \emph{captioned source} video sequences and \emph{uncaptioned target} videos (see illustration in
Figure~\ref{fig:front_page}).
Given the domain gap, pseudo-labelling is noisy. 
Therefore, we propose a robust confidence measure to determine the reliability of pseudo-labels, combined with a sampling strategy that ensures all actions are aligned, rather than only the most confident or common actions.
For better adaptation, we align multi-view embeddings, where each embedding specialises in one part-of-speech (\ie{}~verbs and nouns) within the caption. 

To summarise, our contribution is threefold. First, we introduce the  
problem of Unsupervised Domain Adaptation (UDA) for text-to-video retrieval. In this setup, training only has access to i) a source domain composed of video-caption pairs
and to ii) videos (without captions) from the target gallery set.
Second, we propose a cross-domain learning-to-rank strategy which mixes samples from the source and the target domains thanks to iterative pseudo-labelling and robust sampling strategy. 
Third, we conduct extensive experiments on the EPIC-KITCHENS-100 dataset which validate our approach. Our method outperforms state-of-the-art UDA approaches~\cite{kang2019contrastive,pan2019transferrable,xie2018learning} which we adapted for action retrieval.

\section{Related Work}

\myparagraph{Supervised Cross-modal Video Retrieval}
Learning a joint embedding space is a common approach to perform text-to-video retrieval~\cite{chen2020fine,dong2021dual,gabeur2020multi,hahn2019action2vec,lei2021less,liu2019use,miech2020end,miech2018learning,mithun2018learning,wang2021t2vlad,wray2019fine,zhang2015zero}. Features from pre-trained models have been used as a mixture of experts~\cite{gabeur2020multi,liu2019use,miech2018learning,wang2021t2vlad}. 
Other methods have instead focused on decomposing text captions into 
events, actions and entities~\cite{chen2020fine}, creating a concept space~\cite{dong2021dual},
or disentangling captions into their constituent parts of speech (PoS)~\cite{wray2019fine}.
Additional gains can be achieved by distilling knowledge from vision-text transformers as in~\cite{Miech_2021_CVPR}.
In all these works, no domain gap is expected between the training and test sets.

\myparagraph{Unsupervised Domain Adaptation}
Domain adaptation between pairs of domains (one source and one target), or multiple domains (multi-source), has been proposed for various problems.
Earlier
works acted in the pre-extracted feature space and considered  data re-weighting, metric learning, subspace representations or distribution matching.
Recent methods instead rely on deep architectures, 
trained end-to-end, where the domain discrepancy is minimised jointly with the task error to learn domain invariance~\cite{DamodaranECCV18DeepJDOTOptimalTransportUDA,LongICML15LearningTransferableFeaturesDAN,SunTASKCV16DeepCORALCorrelationAlignment}.
Alternatives to discrepancy minimisation include domain discriminative models with 
adversarial losses~\cite{TzengCVPR17AdversarialDiscriminativeADDA} or
gradient reversal layers~\cite{GaninJMLR16DomainAdversarialNN}.

Aligning the marginal statistics fails to consider the multi-modal structure of input data, ignoring the boundaries between classes and the class imbalance across domains.
Therefore, methods consider the conditional or joint distributions~\cite{LongNIPS18ConditionalAdversarialDomainAdaptation,ZhangICML19BridgingTheoryAlgorithmDA}.
 To approximate target label information,
 soft predictions or pseudo-labels are inferred by a transductive learning paradigm.
Relying on pseudo-labels,
\cite{pmlr-v70-saito17a} designs an asymmetric strategy to learn discriminative
representations for the target domain,   
\cite{chen2019progressive} 
minimises the difference between the class centroids of each domain, while \cite{deng2020rethinking}  enforces a similarity-preserving constraint
 to maintain class-level relations among the source
and target.
To minimise the impact of noisy pseudo-labels during alignment, curriculum learning has been explored~\cite{chen2019progressive,CsurkaTASKCV14DomainDomainSpecificClassMeans,peng2019unsupervised,TommasiICCV13FrustratinglyEasyDA}, where approaches first consider the most confident target samples for alignment, then include the less confident ones at a later stage of the training. 
Confidence scores are determined from classifiers~\cite{peng2019unsupervised,ZhangCVPR18CollaborativeAdversarialUDA}, similarity to neighbours~\cite{SenerNIPS16LearningTransferrableRepresentationsUDA,TommasiICCV13FrustratinglyEasyDA} or to class prototypes~\cite{chen2019progressive,CsurkaTASKCV14DomainDomainSpecificClassMeans}.
To improve the confidence estimation of pseudo-labels,~\cite{roy2019unsupervised} relies on the consensus of image transformations, whereas \cite{pmlr-v70-saito17a}
considers the agreement between multiple classifiers. 

\myparagraph{Domain Adaptation for Videos}
Prior works for video domain adaptation (DA) have focused on classification~\cite{chen2019temporal,ShuffleAttendECCV2020,jamal2018deep,Munro_2020_CVPR}, segmentation~\cite{Chen_2020_WACV, chen2020action} and localisation~\cite{agarwal2020unsupervised}.
They use adversarial training to align the marginal distributions~\cite{jamal2018deep},  an auxiliary self-supervised task~\cite{chen2020action,ShuffleAttendECCV2020,Munro_2020_CVPR}, or attending to relevant frames alignment~\cite{chen2019temporal,Chen_2020_WACV,chen2020action}.

\myparagraph{Domain Adaptation for Retrieval}
Most DA works 
for retrieval were proposed for person re-identification where, 
to overcome the domain shift, 
generative models with similarity preserving constraints learn to transfer the style between domains~\cite{bak2018domain,ChenICCV19InstanceGuidedContextRenderingDAReID,DengCVPR18ImageImageDAwithPreservedSelfSimilarityReID,ge2020structured}. Some methods  exploit 
pseudo-labels obtained by 
joint source and target clustering~\cite{FanTOMM18UnsupervisedPersonReIDClusteringFineTuning,FuICCV19SelfSimilarityGroupingDAReID,ge2020mutual},
while others use curriculum learning to progressively adapt the retrieval space \cite{WuCVPR18ExploitTheUnknownGradually,ZhangICCV19SelfTrainingWithProgressiveAugmentationDAReID}. 

In contrast to 
the above literature, where the retrieval task is performed within a single modality (namely images), 
only few DA works have considered cross-modal retrieval~\cite{Liu_2021_CVPR,peng2019unsupervised}. Initial work~\cite{peng2019unsupervised} learns correlations between captions and images using a scene graph and Maximum Mean Discrepancy (MMD) for domain alignment. 
Concurrent work \cite{Liu_2021_CVPR} also attempts DA for text-to-video retrieval. They train two prototypical classifiers in the cross-modal embedding space, one for the source and one for the target, where labels are found by respectively clustering the source text and target video representations. 
To align domains, 
they maximise the mutual information between the  
prototype assignments of both modalities.
Departing from this approach which uses prototypes for the alignment itself, ours builds on source prototypes to sample a reliable subset of pseudo-labelled target videos for cross-domain learning to rank. We additionally show the benefit of performing alignment in multiple embedding spaces.

\begin{figure*}[t]
    \centering
    \includegraphics[width=1\textwidth]{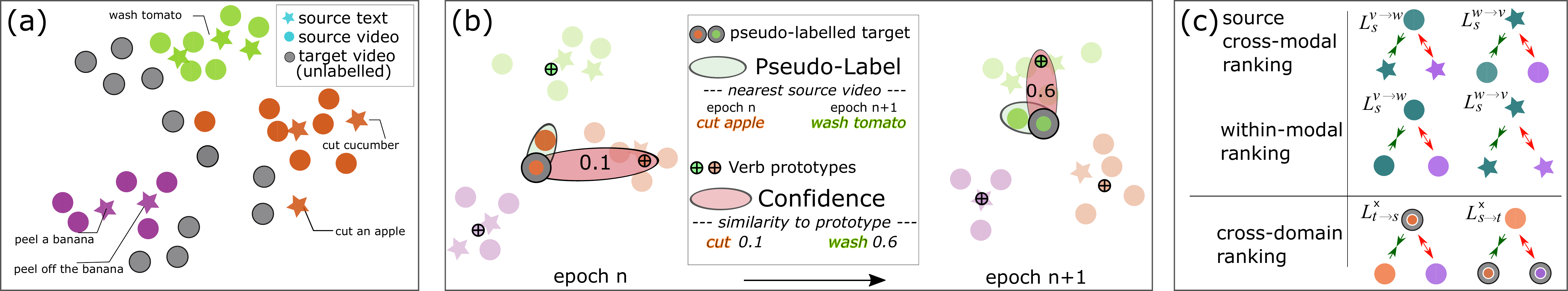}
      \caption{Proposed alignment strategy for the multi-view embedding of verbs. (a) Domain gap between target videos (grey) and video-caption source pairs (coloured according to their verb---wash/green, cut/orange, peel/purple) (b) Target videos inherit pseudo-labels from their nearest source video, with a confidence defined by the verb prototypes. Over epochs, pseudo-labelling is updated. We show one video pseudo-labelled as cut in epoch $n$, and wash in $n+1$ with increased confidence. (c)~Cross-modal losses ($L_s$) are jointly trained with
      cross-domain losses $L^\times_{s \rightarrow t}$, $L^\times_{t \rightarrow s}$. Relevant
      (green arrow) and irrelevant (red arrow) relations are used to train the embedding space with only the most confidently pseudo-labelled target videos.}
    \label{fig:method-figure}
\end{figure*}

\section{Method}
In this work, we tackle text-to-video retrieval, where the goal is to retrieve
relevant videos from a \emph{target} gallery of videos $V^t$, given any text query
$w^t$. To achieve this, we leverage a \emph{source} gallery of paired videos $V^s$ with their captions~$W^s$. 
We consider the case where a domain shift is known to exist between $V^s$ and $V^t$.
Our aim is thus to leverage the paired \emph{source} set, and the single-modality \emph{target} set (\ie videos only), for training a text-to-video retrieval model which will be applied to the 
\emph{target} set.

An overview of our approach is presented in Fig.~\ref{fig:method-figure}.
In this section, we first describe a multi-view text-to-video retrieval
approach which performs well when aligned video-caption
pairs are available for training, \eg as in our source domain~(Sec.~\ref{sec:crossmodal}). Second, we build on this baseline, and present our domain alignment approach~(Sec.~\ref{sec:alignment}) which learns a cross-domain
cross-modal embedding, and uses unsupervised  conditional alignment of the source and target videos along with 
a robust pseudo-labelling strategy applied to the target gallery, which relies on prototype-based confidence estimation~(Sec.~\ref{sec:pseudo}). 

\subsection{Training for text-to-video retrieval with captioned videos}
\label{sec:crossmodal}
 
Given a set of videos $V^s$ and their corresponding captions $W^s$, 
one can learn a joint video-text
embedding space, following a learning-to-rank approach with the 
hinge-based triplet ranking loss. 
Two embedding functions $f: V^s  \rightarrow \Omega$ and  $g: W^s  \rightarrow
\Omega$ which respectively produce representations for videos and text in the embedding space $\Omega$ are learned. The triplet ranking loss is defined as:
   $ H(a,b^+,b^-) = max (\gamma + d\big(h(a), h(b^+)\big) - d\big(h(a), h(b^-)\big), 0)$
where $a$ is the anchor, $b^+$ and $b^-$ are relevant and irrelevant samples for that anchor, 
$\gamma$ is a constant margin, $d(\cdot,\cdot)$ is the distance function and $h \in\{f,g\}$ is
the appropriate embedding function, $f$ or $g$, depending on the modality.

Classical text-to-video retrieval approaches~\cite{chen2020fine,gabeur2020multi,liu2019use,miech2020end} sample triplets within video-caption pairs from $(V^s,W^s)$ and optimize 
    $L(\theta)= \sum H(v_i, w_i, w_j) + \sum H(w_i, v_i, v_k)$, 
where $v_i, v_k \in V^S$ and $w_i, w_j, \in W^S$ and $\theta$ represents the parameters of the embedding functions $h$.
While these approaches typically already perform well, videos/captions of similar actions are not explicitly encouraged to be close in the embedding space.
Inspired by~\cite{wang2016structure,wray2019fine}, we use both cross-modal and within-modal losses between sets of relevant items. 
When the anchor is 
a video,  
combining the cross-modal and
within-modal losses produces: 
\begin{equation}
\begin{split}
      L^{v\rightarrow w}_s(\theta)= \sum H(v_i, w_j, w_k) + \sum H(v_i, v_l, v_m)\\
      \text{with}~w_j \in W^s_{i+}, w_k \in W^s_{i-}, v_l \in V^s_{i+}, v_m \in V^s_{i-}
    \end{split}
    \label{eq:source-losses}
\end{equation}
where $V^s_{i+}$ are all videos relevant to the query video $v_i$
and $V^s_{i-}$ are irrelevant ones. 
Similarly, $W^s_{i+}$/$W^s_{i-}$ is the set of all captions relevant/irrelevant\footnote{We generalise from the case where only a single caption is considered relevant, to a set of relevant captions, in line with recent works~\cite{gordo2017beyond,Wray_2021_CVPR,zhou2019ladder}. Finding these sets of relevant videos/captions requires additional knowledge, such as visual clustering~\cite{FuICCV19SelfSimilarityGroupingDAReID}, 
attributes~\cite{ak2018learning}, or semantic similarity between captions~\cite{gordo2017beyond,Wray_2021_CVPR,zhou2019ladder}. 
}  
to the query video $v_i$.

Analogous to Eq.~\ref{eq:source-losses}, text-to-video cross-modal and within-modal losses are defined,  where the anchor is a caption, as $L^{w\rightarrow v}_s$. The training is then defined as 
$L_s= L^{v\rightarrow w}_s+L^{w\rightarrow v}_s$.

\myparagraph{Multi-view Embeddings}
Recent works in text-to-video retrieval have shown that learning multiple embedding spaces (or multi-view embeddings) could be beneficial~\cite{gabeur2020multi,liu2019use,miech2018learning,wray2019fine}. This can be achieved by disentangling the visual input into multiple modalities~\cite{liu2019use,miech2018learning} or the caption into individual parts of speech~\cite{wray2019fine}.
Due to our focus on fine-grained actions, we follow the JPoSE architecture~\cite{wray2019fine}: 
We parse captions into parts of speech (PoS) and we train a separate embedding for each.
The embedding vectors in the various views are then concatenated, and considered as input to a final action embedding, where text-to-video retrieval can be performed.
Each embedding is trained using its own set of triplets involving embedding-specific relevance relations. In the `verb' embedding, all captions involving the verb (\eg~`cut') are considered relevant to one another, regardless of the object.
Conversely, in the `noun' embedding, all captions involving the same noun are considered relevant. 

In the following, we  show how a joint multi-view embedding space trained on the source domain ${S = (V^s, W^s)}$ can be adapted for the target video gallery $V^t$. The proposed adaptation method is applied to \emph{each} of the multi-view embeddings, trained jointly. 

\subsection{Cross-domain Cross-modal Embedding}
\label{sec:alignment}
\label{sec:cross-domain-embedd}

Reducing the domain shift by aligning marginal
distributions of the source and the target video sequences
often leads to sub-optimal solutions as it does not consider the alignment of local distributions between the two galleries.
We propose a cross-domain cross-modal approach instead.
To overcome the lack of captions in
the target gallery $V^t$ 
we instead propose using pseudo-labels to model cross-modal relevance.

For simplicity, we refer to source video $v^s_i \in V^s$ as $s_i$, and to  target video $v^t_i \in V^t$ as $t_i$.
We define the source-to-target and 
target-to-source 
cross-domain ranking objectives as:
\begin{equation}
\begin{split}
      L^\times_{s \rightarrow t}(\theta)=\sum
      H(s_i, t_j, t_k)
      |~
      t_j \in V^t_{s_i+}, t_k \in V^t_{s_i-}\\
      L^\times_{t \rightarrow s}(\theta)=\sum
      H(t_i, s_j, s_k)
     |~
      s_j \in V^s_{t_i+}, s_k \in V^s_{t_i-}
      \end{split}
    \label{eq:cross-domain}
\end{equation}
where $V^t_{s_i+}/V^t_{s_i-}$ are  relevant/irrelevant sets of target videos for
the source video $s_i$ and
$V^s_{t_i+}/V^s_{t_i-}$ are the relevant/irrelevant sets of source videos for
the target video $t_i$.
These relevant sets are defined by the pseudo-labelling strategy described in Sec~\ref{sec:pseudo}.

We then train the 
source embedding  objectives 
(see Section \ref{sec:crossmodal})
along with the cross-domain ranking objectives, \ie 
    $L = L_s + \lambda_1 L^\times_{s \rightarrow t} + \lambda_2 L^\times_{t \rightarrow s} $
where $\lambda_1$ and $\lambda_2$ are weights to balance the corresponding cross-domain losses.

Importantly, note that this is an iterative process during training. As the joint embedding space is trained, 
target videos are pseudo-labelled, and are used to generate triplets for cross-domain ranking.
Target videos can potentially be assigned to different pseudo-labels at each iteration. 
For robust learning, we adjust 
the assignments to the relevance sets
at the end of each training epoch. 

As mentioned in the previous section, we learn 
a separate embedding space for each part of speech together with a final action embedding. We apply Eq.~(\ref{eq:cross-domain}) to each PoS embedding (\ie view embedding) as well as to the overall action embedding. 

\subsection{Pseudo-labelling and Prototype-based Sampling}
\label{sec:pseudo}

In order to sample triplets for the cross-domain losses in Eq.~(\ref{eq:cross-domain}), 
we need pairs of relevant videos between the source and the target sets.
Therefore, we propose: 
i) a pseudo-labelling strategy to determine relevant video pairs across domains, and 
ii) an associated prototype-based confidence measure 
to select which target videos to use for training.

In standard DA, which often assumes a single modality and a
classification task, typical approaches use the softmax scores of the 
source classifier to produce pseudo-labels~\cite{pmlr-v70-saito17a,DengICCV19ClusterAlignmentTeacherUDA},
or  label propagation on the nearest neighbour graph built with the joint source and target sets~\cite{SenerNIPS16LearningTransferrableRepresentationsUDA,iscen2019label}.
Taking inspiration from these strategies,
and replacing 
{\em pseudo-labelling} with the process of assigning to each target video a relevant set of source videos,  we proceed as follows.
As in classical label propagation, the target video inherits the relevance property of the closest
source video, \ie a relevant/irrelevant query to the nearest source video is considered relevant/irrelevant to the target video. Formally, if $s_{\hat{n}}$ is the nearest source video to the target video, $t_i$, 
\ie ${\hat{n} = \arg\min_n d(f(t_i),f(s_n))}$,
we assign 
$        V^s_{t_i+} = \{V_{s_{\hat{n}}+}^s \cup \{s_{\hat{n}}\}\}; 
        V^s_{t_i-} = V_{s_{{\hat{n}}-}}^s.
$
Similarly, the target video is considered relevant to each source video in its relevant set.
However, this process which associates 
videos across the source and the target domains 
 might be error prone, and relying on too many erroneous relevance pairs would degrade the model. On the other hand, 
 trusting too few would not be sufficient to align the domains.
 Consequently, to handle such a trade-off, we propose 
to robustly measure the confidence of the labelling process
and to use these confidence scores to select target videos. 

\myparagraph{Prototype Confidence (PC)} We propose to calculate confidence scores based on the distance between the target videos and the \emph{source prototypes} of their closest source video $s_{\hat{n}}$ in the embedding space. 
A  source {\em prototype} is defined as the barycentre of the sets of relevant videos as described in Sec.~\ref{sec:cross-domain-embedd}. All videos within one set $V^s_{s_i+}$ are relevant to each other and share the same relevance properties, \eg they show the same action/PoS. We calculate the
 prototype for this set $V^s_{p}$ by 
 $\mu_p=\sum_{s_i\in V^s_{p}} f(s_i)$.
 
Let $\mu_{\hat{p}}$ be the prototype corresponding to  the source video $s_{\hat{n}}$, \ie $s_{\hat{n}} \in V^s_{\hat{p}}$, and therefore  $V^s_{s_{\hat{n}}+}=V^s_{\hat{p}}$. Then, the 
  confidence measure  for pseudo-labelling the target video, $t_i$,
  is calculated as a function of the distance to the prototype  $\mu_{\hat{p}}$, more precisely 
  $\alpha_{t_i} = e^{- d(f(t_i), \mu_{\hat{p}})}$.
 
 Note that we propose to calculate the distance between the target video and the closest prototype $\mu_{\hat{p}}$, rather than the closest source video $s_{\hat{n}}$, so we increase robustness by avoiding outliers  and considering the distance of the target video to all relevant source videos. 

Importantly, the embeddings $f(t_i)$, $f(s_{\hat{n}})$ and prototypes $\mu_{\hat{p}}$ depend on the learnt embedding function $f$, and hence they continuously
change over training iterations. Subsequently, the pseudo-labelling and the confidence scores are also continuously changing, but it would be extremely costly to update them 
for all target videos
after every model update. Fortunately, this change is progressive and updating them at the end of every epoch is sufficient. 
   
\myparagraph{Prototype Based Sampling (PBS)} Using the proposed prototype confidence scores, we can sample the most reliable target videos, and use them for training our cross-domain losses~(Eq.~\ref{eq:cross-domain}).
However, selected target videos do not only need to be reliable, they also need to widely cover the different actions. 
Without sufficient coverage, some actions can be poorly aligned.
As training progresses, the variety of selected actions may decrease as the labelling gets biased towards actions that are already aligned.

To avoid only aligning a few confident actions, 
and ignoring the others during domain alignment,
we leverage the source prototypes.
Let us consider the set of all target videos that are pseudo-labelled using the source prototype $p$, \ie their relevant source videos are $V^s_{p}$. 
We would like to select the most confident of these, per prototype, to maintain coverage.
  We thus propose to sample the top $x\%$ most confident target videos assigned to each prototype, 
  \ie  we rank the videos assigned to $V^t_{p}$ 
   based on the confidence scores $\alpha_{t_i}$ 
   and take the top $x\%$ of the list, for each prototype.
   Our proposal thus can deal with imbalanced sets, maintaining the distribution of pseudo-labels, when sampling the most confident ones.

To summarise, we align each multi-view embedding (Sec~\ref{sec:crossmodal}) by pseudo-labelling, sampling the most confident target videos per prototype, and iterating.
Importantly, the multi-view embeddings are trained jointly, with one loss $L$ per embedding.

\section{Experimental Evaluation}

\myparagraph{Dataset}
We evaluate our approach on EPIC-KITCHENS-100~\cite{Damen2020RESCALING}, which provides videos associated with free-form narrated captions of egocentric actions. 
The Source/Target split has only been evaluated in the context of action recognition~\cite{Munro_2020_CVPR}, using class labels but not the captions.

We observe a similar visual domain gap to that in action recognition (see test performances in Fig.~\ref{fig:front_page}).
We consider all videos from the TRAIN/VAL UDA action recognition challenge, as these contain released captions\footnote{The Test split in the challenge does not have released captions publicly available}.
They are collected from 16 participants, with 
videos recorded with a two-year gap~\cite{Damen2020RESCALING}, totalling 55K action videos (21K Source  and 34K Target).
The domain gap in this dataset 
results from recording footage in different environments and recording periods.

\begin{table}[t]
    \centering
    \small
    \setlength\extrarowheight{-3pt}
    \begin{tabular}{llccc}
    \toprule
        Split &Domain &Participants &Video Gallery &Text Queries\\ 
        \midrule
       TRAIN &Source / Target &12 / 12 &16115 / 26115 & 4756 / [5907$^\dagger$] \\
       
        \midrule
         VAL &Source / Target &4 / 4 &5002 /7906 & 1805 / [2822$^*$]  \\
        \bottomrule
    \end{tabular}
    \vspace{0.5cm}
    \caption{Proposed EPIC-KITCHENS benchmark. 
    Note that target text queries are \textbf{never} used for training. $\dagger$: only used for evaluation *: only used for optimising hyper-parameters.}
    \label{tab:splits}
\end{table}

\myparagraph{Proposed Benchmark} We propose the first benchmark for domain alignment in fine-grained text-to-video retrieval (see Table~\ref{tab:splits}). 
We use the VAL split of the UDA challenge~\cite{Damen2020RESCALING} only to select  the hyperparameters.
Once hyperparameters are decided, we train the model on the TRAIN split. 
The model is trained with Source videos and their captions but only Target videos are used (with no captions). At test time, the model is evaluated by ranking the Target videos for each Target text query. 
For evaluation, we report mAP and nDCG, both of which measure ranking quality, as proposed in the supervised retrieval challenge~\cite{Damen2020RESCALING}. 
We follow the same approach to define ground-truth retrieval results, considering retrieved videos depicting the same actions as described by the textual query as relevant. 

\myparagraph{Model and Pretraining}
For our experiments we use the public implementation of JPoSE~\cite{wray2019fine} as our base network and modify it for domain adaptation as described in Sec.~\ref{sec:cross-domain-embedd} and~\ref{sec:pseudo}.
To train the model, we first pre-train our network only on Source and use it to initialise the proposed domain alignment network.
See supplementary for further details.

\myparagraph{Alignment Baselines}
As a lower bound, we
include the non adapted \textbf{Source-Only} results. 
Additionally, we implement per-domain standardisation \textbf{PDS}
which is a simple alignment technique where the input distribution of each domain is standardised separately. PDS decreases the domain shift. All methods---including ours---are trained with these standardised features.
We compare our method to
classification-based (\ie typical) domain alignment methods, that we modified for the task of cross-modal video retrieval.  All methods were hence trained and tested with the exact same protocol. We list them below. 

From shallow models we test \textbf{CORAL}~\cite{coralAAAI2016}, which 
aligns second order statistics, and from deep
models we consider \textbf{MMD}~\cite{LongICML15LearningTransferableFeaturesDAN},  which minimises the discrepancy 
between source
and target embeddings, and  \textbf{GRL}~\cite{GaninJMLR16DomainAdversarialNN}, which optimizes a domain discrimination loss in an adversarial manner. We also compare to more recent conditional alignment approaches, adapting these 
by replacing the
notion of classes with groups of relevant videos that share the same prototype. In particular, we evaluate 
\textbf{MSTN}~\cite{xie2018learning} which minimizes the Euclidean distance between the source and target prototypes, 
\textbf{TPN}~\cite{pan2019transferrable} which minimises the MMD between the source and target examples per prototype and 
\textbf{CDD}~\cite{kang2019contrastive} where the MMD is minimised within 
each prototype while maximised
across examples belonging to differing prototypes.  
For fairness, all conditional alignment methods use the same pseudo-labelling strategy as our method and, to make sure conditional alignment baselines have a sufficient number of instances per prototype, we compare alignment of the Part of Speech (PoS) embeddings. 

\begin{figure}[t]
\small
\begin{minipage}{\textwidth}
        \centering
        \setlength\extrarowheight{-5pt}
        \begin{tabular}{llrccccllll}
        \toprule
          & Method  & \multicolumn{2}{c}{Alignment Space} & \multicolumn{2}{c}{} & \multicolumn{2}{c}{Val (Target VAL)}  & \multicolumn{2}{c}{ Test (Target TRAIN)}  \\
          & & PoS & Action &M &C &nDCG & mAP & nDCG & mAP \\
          \midrule
          \multirow{8}{*}{(a)} & Source-Only & $\times$ & $\times$ & $\times$ & $\times$ & $40.45$ & 7.32 & $35.25$ & 5.05\\
          \cmidrule(l){2-10}
          & PDS & $\checkmark$ & $\times$ & $\checkmark$ & $\times$& $40.78$& 7.65 & $35.92$ & 5.42\\
          \cmidrule(l){2-10}
          & CORAL~\cite{coralAAAI2016} & $\checkmark$ & $\times$ & $\checkmark$ & $\times$ & $40.70$ & 7.79& $36.32$ & 5.38\\
          \cmidrule(l){2-10}
          & MMD~\cite{LongICML15LearningTransferableFeaturesDAN} & $\checkmark$ & $\times$ & $\checkmark$ & $\times$ & $40.60$ & 7.49& $36.26$ & 5.43\\
          & GRL~\cite{GaninJMLR16DomainAdversarialNN} & $\checkmark$ & $\times$ & $\checkmark$ & $\times$ &$42.45$ & 5.20 & $36.64$ & 5.61\\ \cmidrule(l){2-10}
          & TPN~\cite{pan2019transferrable} & $\checkmark$ & $\times$ & $\checkmark$ & $\checkmark$ & $42.19$ & 8.27 & $36.86$ & 5.73\\
          & CDD~\cite{kang2019contrastive} & $\checkmark$ & $\times$ & $\checkmark$ & $\checkmark$  & $42.00$ & 8.48 & $36.48$ & 5.80\\
          & MSTN~\cite{xie2018learning} & $\checkmark$ & $\times$ & $\checkmark$ & $\checkmark$  & $42.85$ & \textbf{8.63}
          & $37.62$ & 5.87\\
          \cmidrule(l){2-10}
          & Ours-PoS & $\checkmark$ & $\times$ & $\checkmark$ & $\checkmark$ & \textbf{43.14} & 8.59 & \textbf{38.01} & \textbf{6.21} \\
          \bottomrule
          \toprule
          \multirow{2}{*}{(b)}           
          &Ours-Action & $\times$ & $\checkmark$ & $\checkmark$ & $\checkmark$ & 43.07 & 8.85 & 37.74 &  6.20\\

          & Ours  & $\checkmark$ & $\checkmark$ & $\checkmark$ & $\checkmark$ & \textbf{44.00} & \textbf{9.10} & \textbf{38.21} &  \textbf{6.34}\\
          \bottomrule
        \end{tabular}
        \captionof{table}{(a) Comparison with alignment baselines (M: Marginal alignment, C: Conditional alignment). TPN, CDD, MSTN use our sampling strategy for fair comparison. (b) Which space(s) to use for alignment: Part-of-Speech or Action (see sec.~\ref{sec:crossmodal}).} 
        \label{tab:comparative_results}
\end{minipage}
\end{figure}


\myparagraph{Comparison to SOTA} Table~\ref{tab:comparative_results}(a) shows how the proposed method performs against models trained solely on source (Source-Only) and both marginal and conditional alignment baselines. 
We report retrieval results on Target VAL (referred as Val) and on Target TRAIN (referred as Test\footnote{We report the best Val epoch based on nDCG  and  use it as early stopping criteria when testing on Target TRAIN.}). 
Surprisingly, the simpler marginal alignment methods (PDS and CORAL) yield better performance than the deep models (MMD) for our benchmark, but overall they provide marginal improvements over the Source-Only model. Adversarial training (GRL) provides a larger improvement, however, we find this to be highly unstable as training progresses. The conditional alignment approaches (TPN, CDD, MSTN) perform better than marginal alignment, with MSTN performing the best.
Our approach (Ours-PoS), using cross-domain alignment, outperforms all baselines on Test.

 In Table~\ref{tab:comparative_results}(b) we show two alternatives of the proposed model, Ours-Action where we align the domains in the Action embedding space.
 Even with single-view domain alignment (only Action embedding), our approach outperforms all baselines. 
Finally, by aligning all embeddings (PoS and Action) jointly, we observe further improvements over Ours-PoS. 

To verify the design choices of our proposed model (see Sec.~\ref{sec:pseudo}) we answer the following three questions in Table \ref{tab:ablation} (further details in the supplementary): 

\myparagraphbase{\em How to pseudo-label?} 
Table~\ref{subtab:labelling} shows that 
inheriting from the nearest source video performs better than 
using the nearest prototype's label (Proto) because it makes fewer assumptions on the target distribution.

\myparagraphbase{\em How to assign confidence?} 
Table~\ref{subtab:confidence} shows that
 the proposed confidence measure $\alpha$---
based on the distance to the prototype---
is more robust than using a confidence based on the distance to the closest source video instead 
(Neighbour). 

\begin{wrapfigure}[12]{r}{0.47\textwidth}
    \begin{minipage}{0.47\textwidth}
    \vspace{-0.5cm}
    \includegraphics[width=0.9\textwidth]{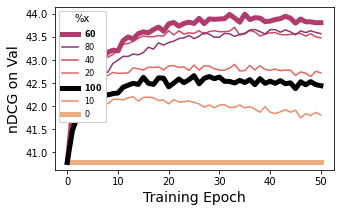}
    \captionof{figure}{Performance as we train for different proportions of target sampled.} 
    \label{fig:percent-target}
    \end{minipage}
    \begin{minipage}{0.49\textwidth}
    \end{minipage}
    
\end{wrapfigure}

\myparagraphbase{\em How to sample videos?} 
In Table~\ref{subtab:sampling} we compare the proposed sampling strategy to uniform sampling. 

By sampling confident examples per prototype, we ensure that they cover a variety of actions, thus improving the domain alignment. Without our sampling strategy, target videos are sampled from fewer actions as training progresses. 

\myparagraphbase{\em How many relevant?}
Figure~\ref{fig:percent-target} shows the impact of different proportions used to sample target examples. 
The results are shown on Val.
We observe that removing the least confident examples significantly improves target performance and that $60\%$ performs best. Also, we see that, as expected, using all target instances (100\%), which would include erroneous pseudo-labels, leads to a substantial drop in performance. This validates our sampling proposal.

\begin{table}[t!]
    \small
    \centering
    \begin{subtable}{.3\linewidth}
    \centering
    \caption{Labelling Method}
    \begin{tabular}{lr}
    \toprule
    Method & Test   \\ \midrule
    Proto & $37.92 \pm 0.12$   \\
    Ours  & $38.21 \pm 0.08$  \\
    \bottomrule
    \label{subtab:labelling}
    \end{tabular}
    \end{subtable}
    \begin{subtable}{.3\linewidth}
    \centering
    \caption{Confidence Method}
    \begin{tabular}{lr}
    \toprule
    Method & Test     \\ \midrule
    Neighbour    &  $37.64 \pm 0.30$   \\
    Ours &  $38.21 \pm 0.08$   \\
    \bottomrule
    \label{subtab:confidence}
    \end{tabular}
    \end{subtable}
    \begin{subtable}{.38\linewidth}
    \centering
    \caption{Video Sampling Method}
    \begin{tabular}{lr}
    \toprule
    Method          & Test     \\ \midrule
    Uniform $(60\%)$ &  $37.88 \pm 0.08$   \\
    Ours $(60\%)$  &  $38.21 \pm 0.08$   \\
    \bottomrule
    \label{subtab:sampling}
    \end{tabular}
    \end{subtable}
    \caption{Ablation Studies on the Test set (nDCG). Mean and standard deviation over 3 runs.
    \label{tab:ablation}}
\end{table}

\section{Conclusion}
In this work, we have introduced the problem of unsupervised domain adaption for text-to-video retrieval. Given a source dataset of captioned videos for training, the goal is to perform retrieval on a distinct target set of uncaptioned videos. 
At the heart of our proposed approach lies an iterative pseudo-labelling process, which associates source and target samples at train time. These pseudo-labelled samples fuel our cross-domain cross-modal learning-to-rank approach which aligns the source and the target video galleries. 
Our experiments validate this strategy and highlight the importance of selecting confident target examples, using source prototypes, during alignment. 

\appendix
\section*{Appendix}
In this appendix we provide additional analysis and further implementation details. We show qualitative results of text-video retrieval using target text queries in Appendix~\ref{sec:qualitative} followed by feature visualisations in Appendix~\ref{sec:feature_vis}. Next we show the accuracy of pseudo-label assignments in Appendix~\ref{sec:pseudo}, and the variety of pseudo-labels assigned to target videos in Appendix~\ref{sec:variety}.
Finally, we report the Mean Average Precision metric for the ablation in the main paper (Table \ref{tab:ablation}) in Appendix~\ref{sec:map}, and discuss the implementation details in more depth in Appendix~\ref{sec:furtherimplementation}. 
\section{Qualitative Results}
\label{sec:qualitative}
\begin{figure*}[hhh]
    \includegraphics[width=\linewidth]{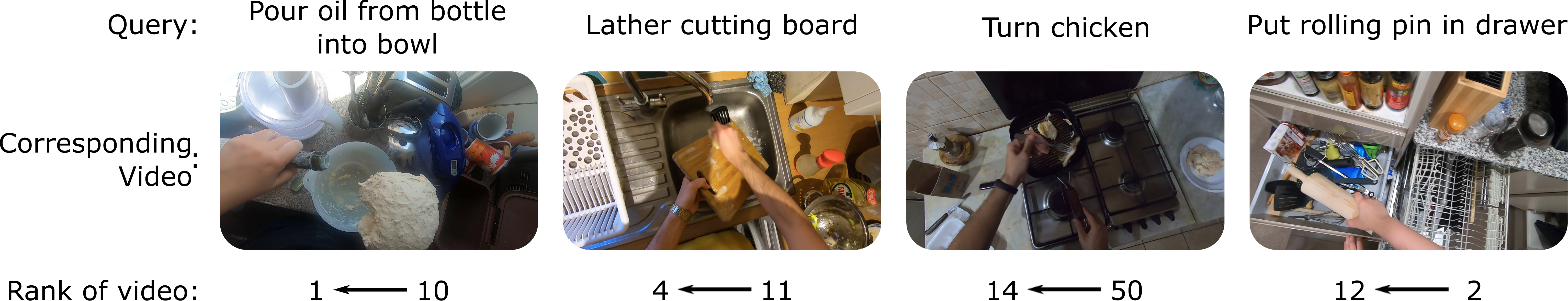}
    \caption{Qualitative results of text-to-video retrieval. We show the query caption and the corresponding video along with the change in rank, $A \leftarrow B$, of the proposed method (A) and Source-Only baseline (B).}
    \label{fig:qualitative}
\end{figure*}

Figure~\ref{fig:qualitative} shows how the rank of the first relevant video to specified text queries changes from the Source-Only model and Ours. Ours results in more relevant videos retrieved higher in the ranking, however, some failure cases exist. The video of "put rolling pin in drawer" is retrieved lower in the ranking, with  videos depicting actions of moving utensils retrieved higher in the ranking. 

\section{Feature Visualisations}
\label{sec:feature_vis}
\begin{figure}
    \centering
    \textbf{Analysis of the alignment of source and target videos in the action and part-of-speech embeddings}
    \begin{minipage}{\linewidth}
    \includegraphics[width=0.49\textwidth,frame]{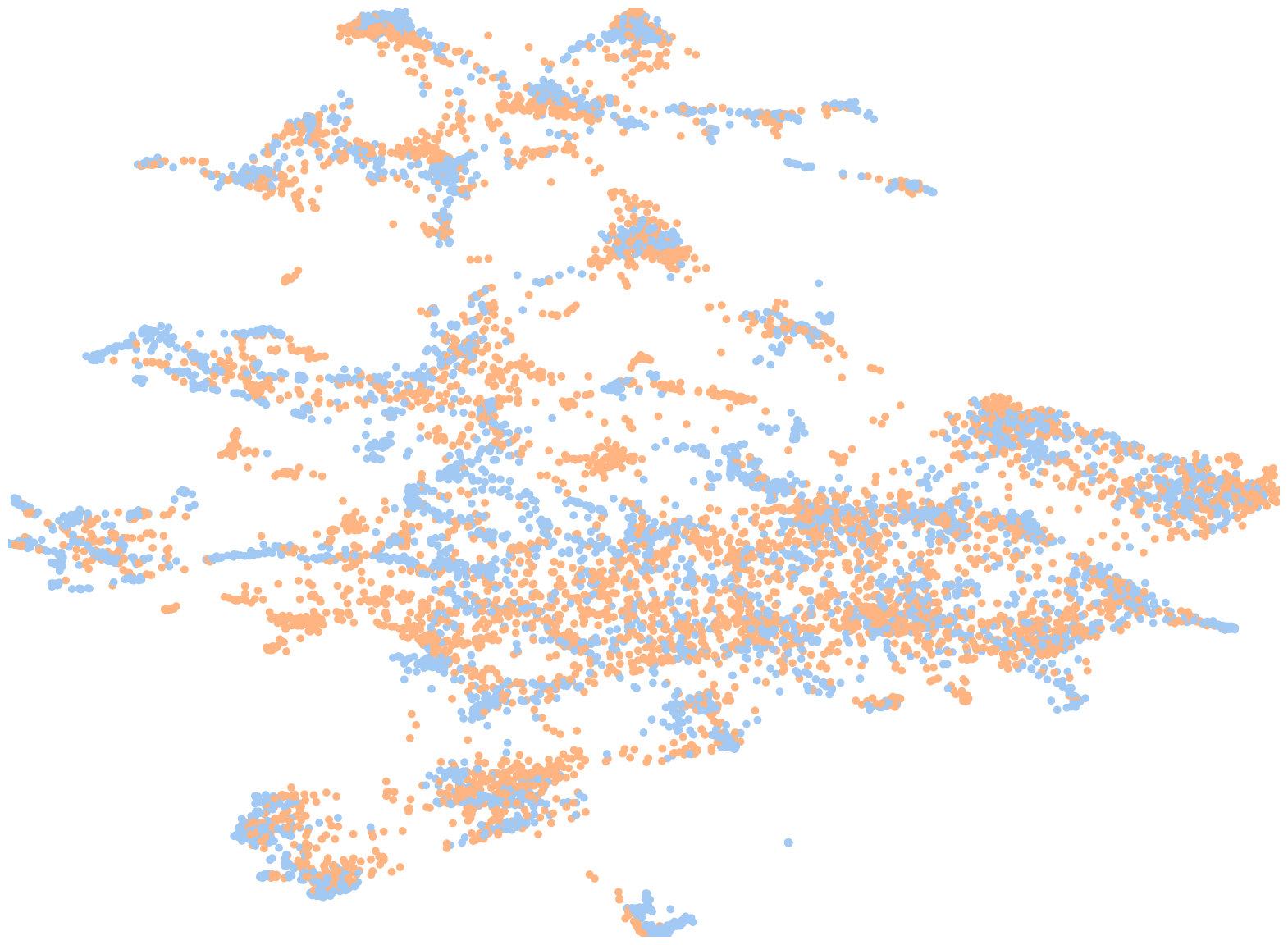}
    \includegraphics[width=0.49\textwidth,frame]{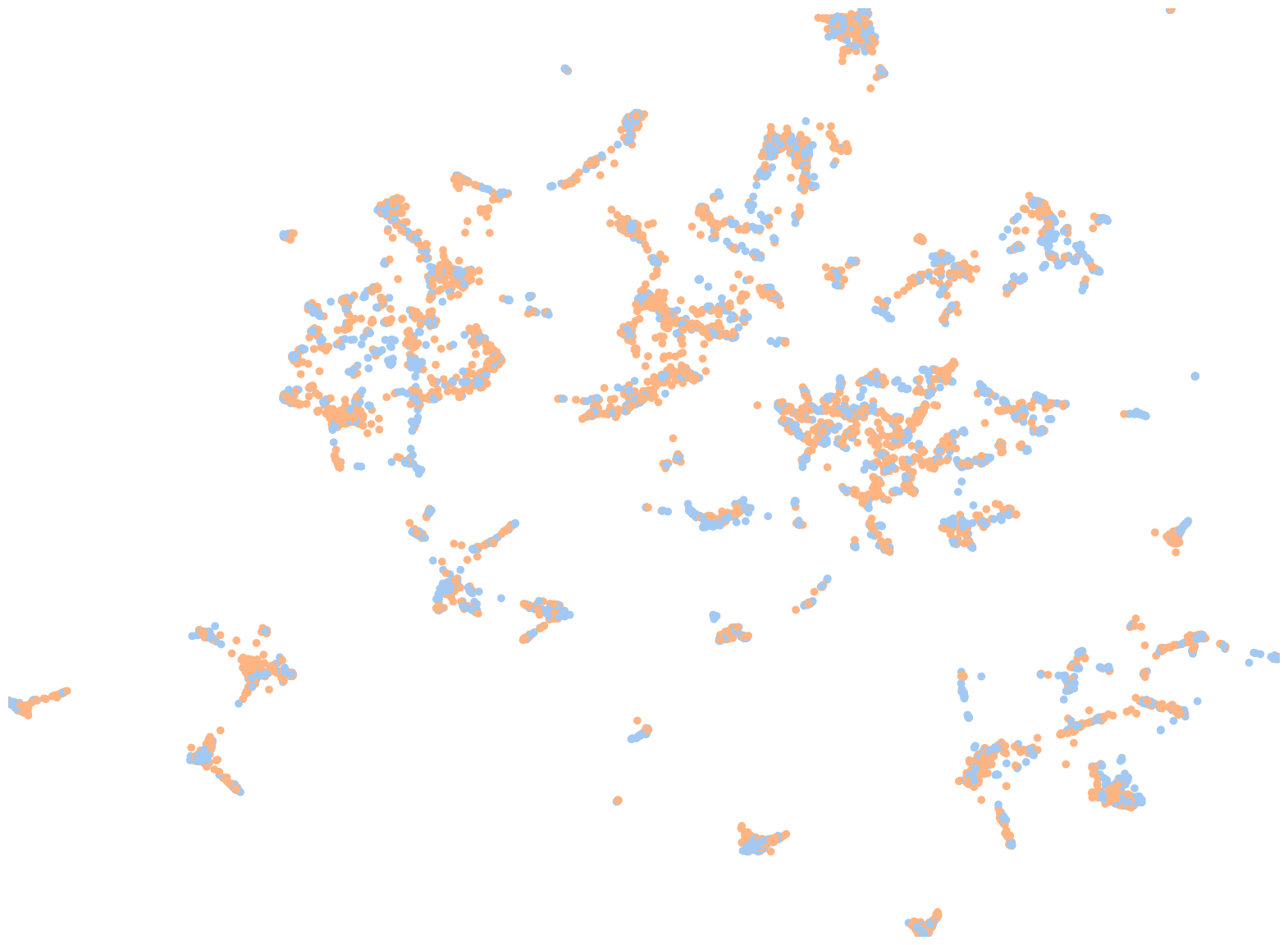}
    \subcaption{Action Embedding}
    \end{minipage}
    \begin{minipage}{\linewidth}
    \includegraphics[width=0.49\textwidth,frame]{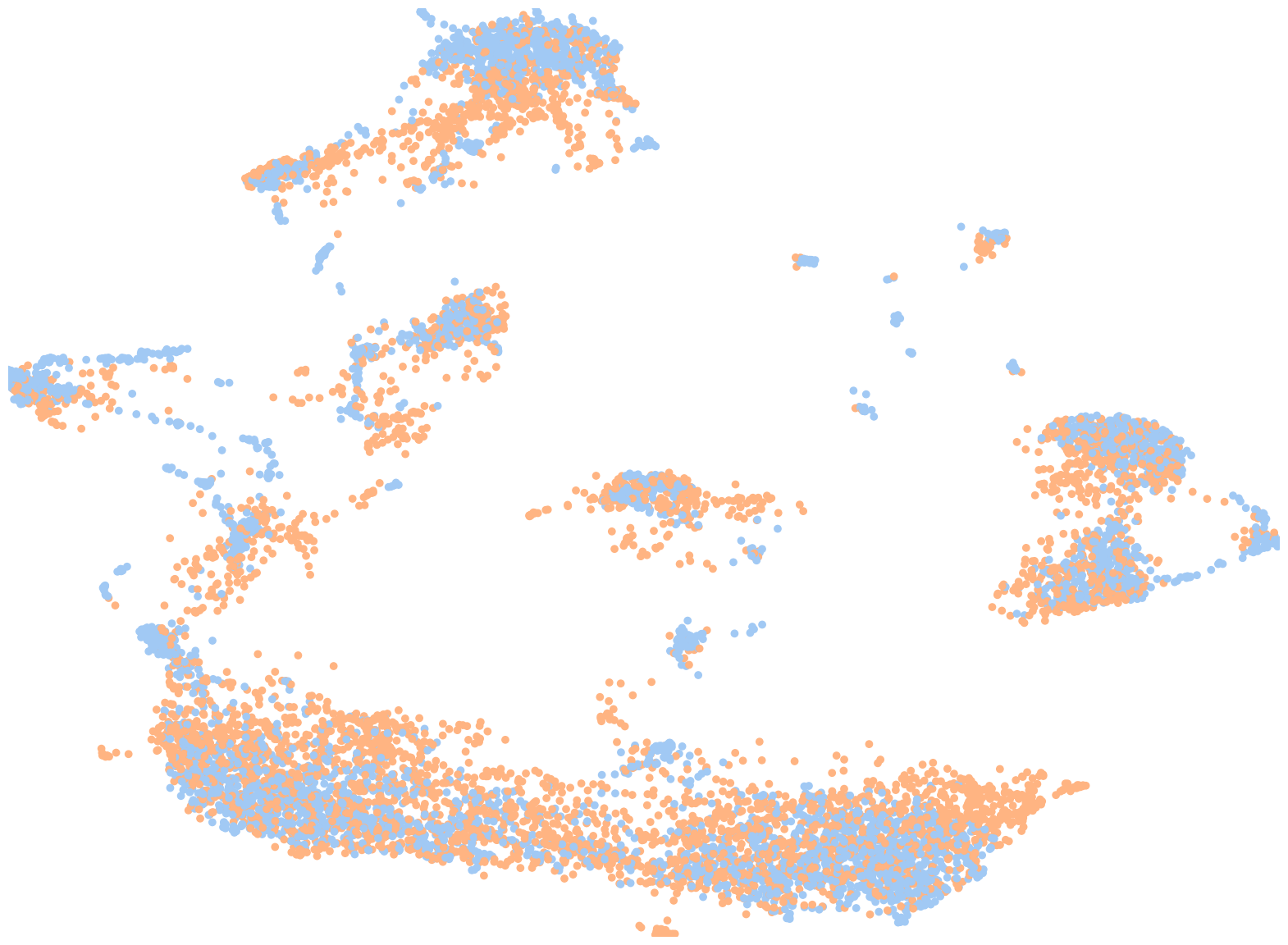}
    \includegraphics[width=0.49\textwidth,frame]{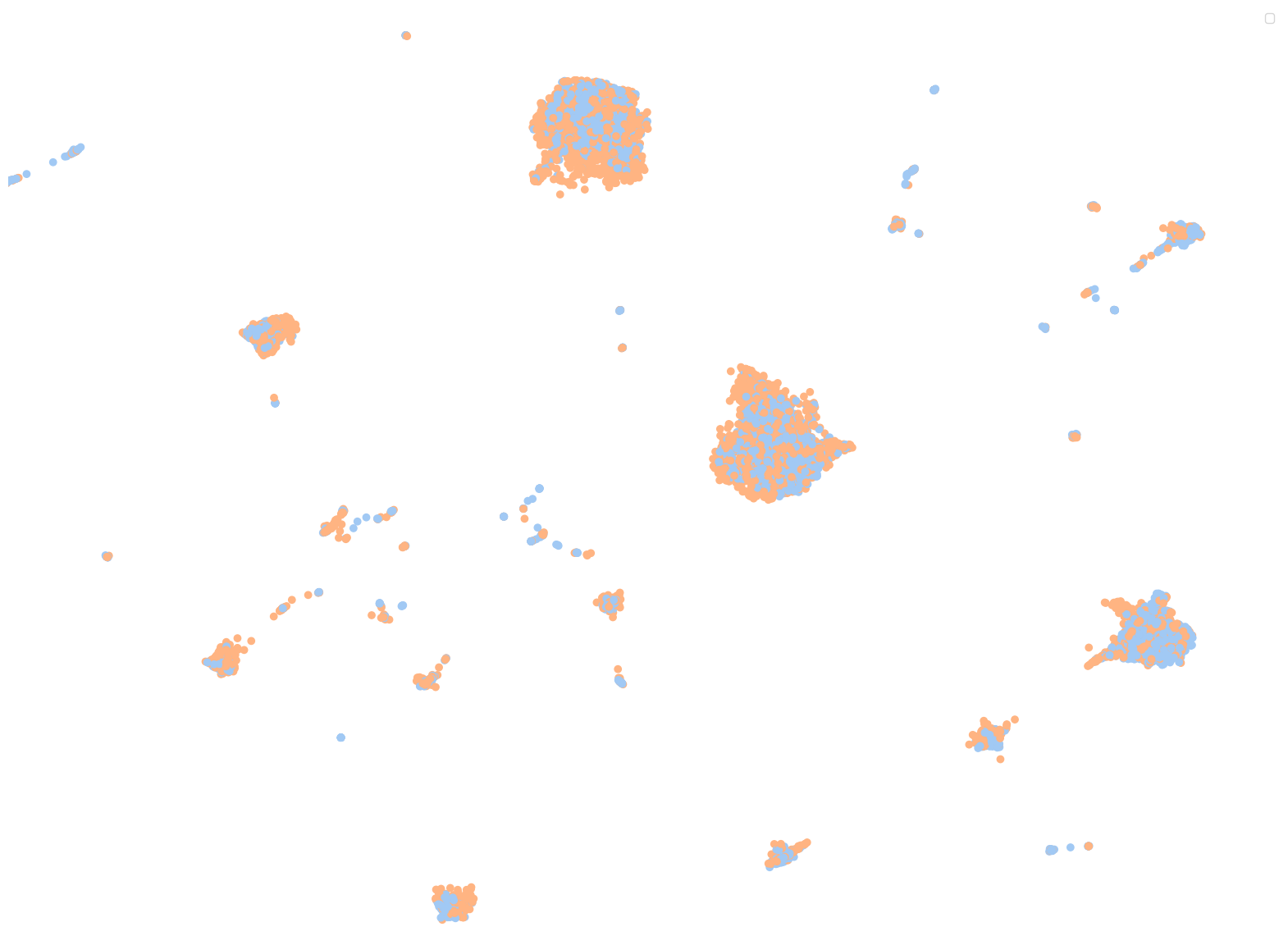}
    \subcaption{Verb Embedding}
    \end{minipage}
    \begin{minipage}{\linewidth}
    \includegraphics[width=0.49\textwidth,frame]{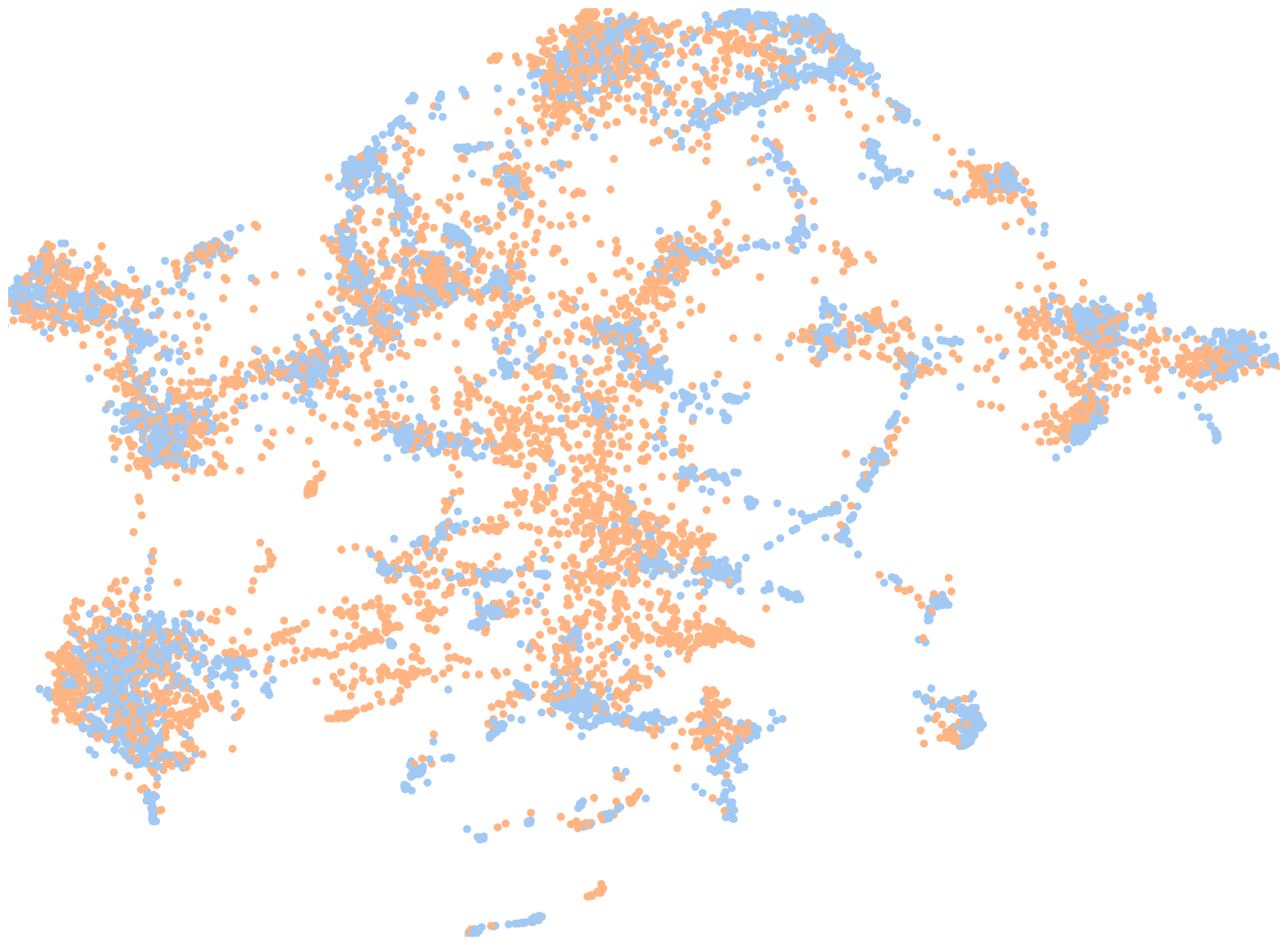}
    \includegraphics[width=0.49\textwidth,frame]{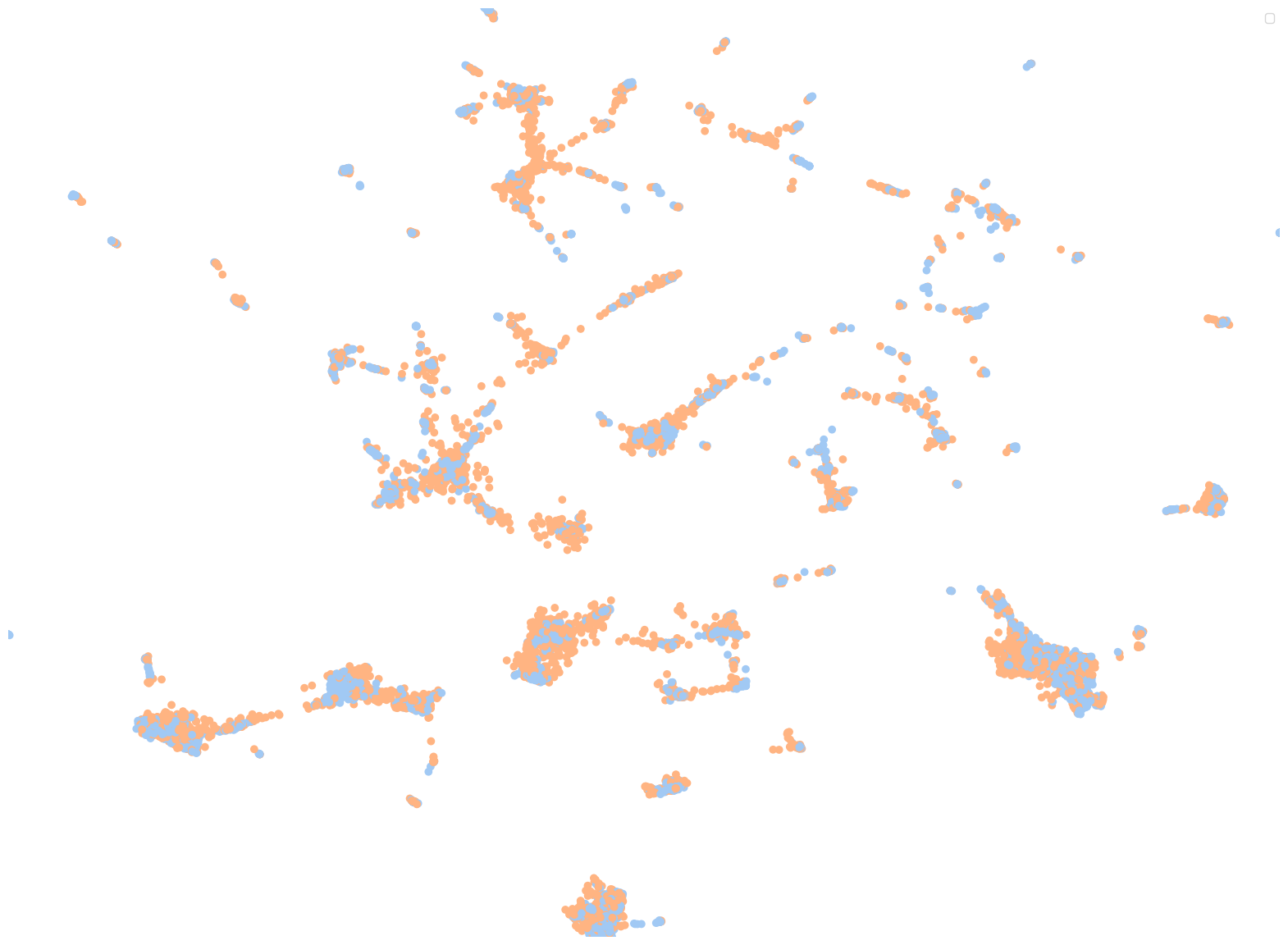}
    \subcaption{Noun Embedding}
    \end{minipage}
    \vspace{1mm}
    \caption{UMAP visualisation of the part-of-speech and action embeddings, Target in ORANGE and Source in BLUE.  LEFT: Source-only, RIGHT: Ours}
    \label{fig:part-of-speech-vis}
\end{figure}

Fig~\ref{fig:part-of-speech-vis} shows the UMAP visualisation of the action, noun and verb embedding spaces for Source-Only (LEFT) and Ours (RIGHT). Ours not only shows greater alignment of source (BLUE) and target (ORANGE) videos but more distinct clusters of videos depicting the same action/part-of-speech. 

\begin{figure}
    \centering
    \textbf{Analysis of the alignment of target text queries with target videos}
    \begin{minipage}{\linewidth}
    \includegraphics[width=0.9\textwidth,frame]{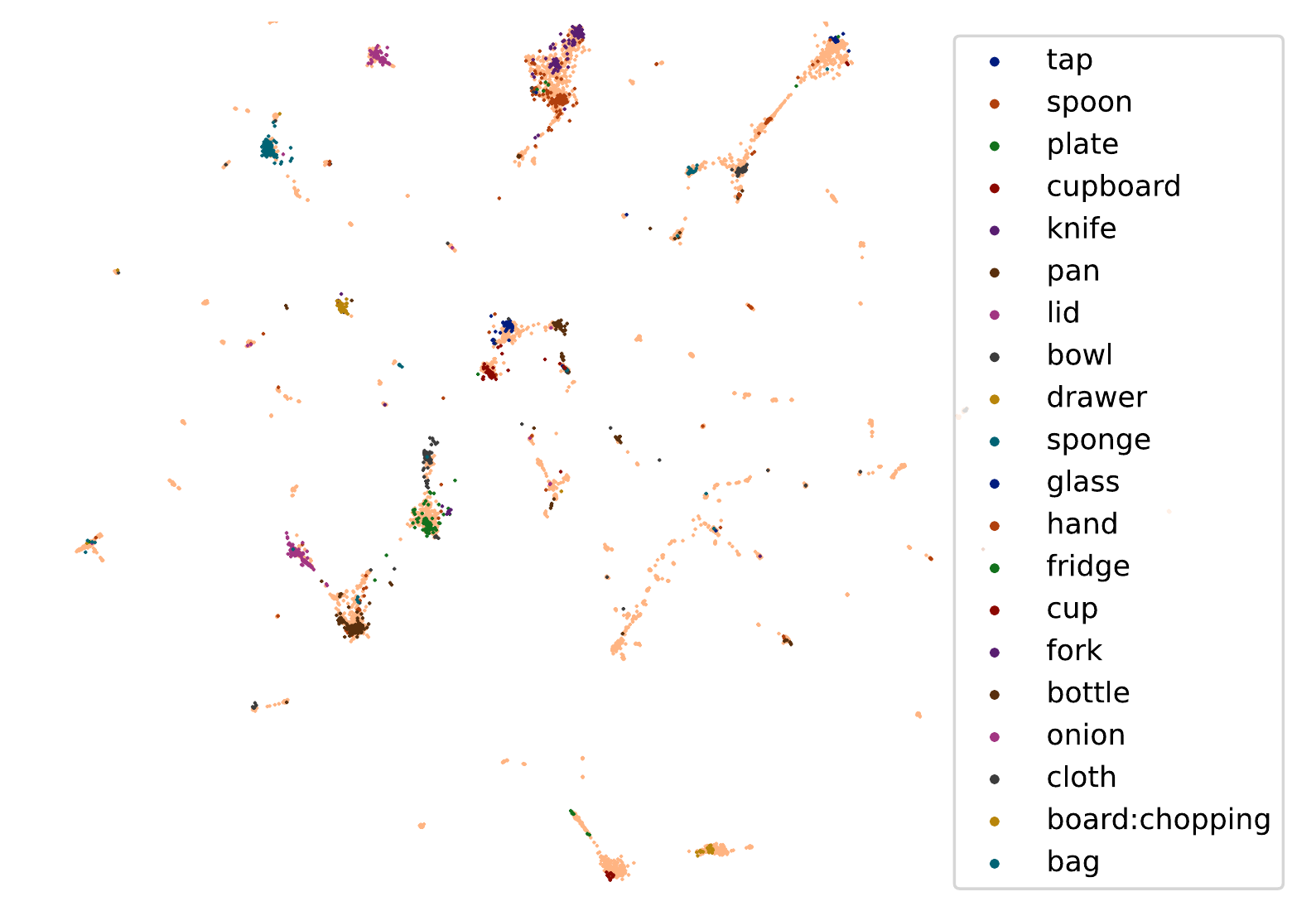}
    \subcaption{Noun embedding}
    \end{minipage}
    \begin{minipage}{\linewidth}
    \includegraphics[width=0.9\textwidth,frame]{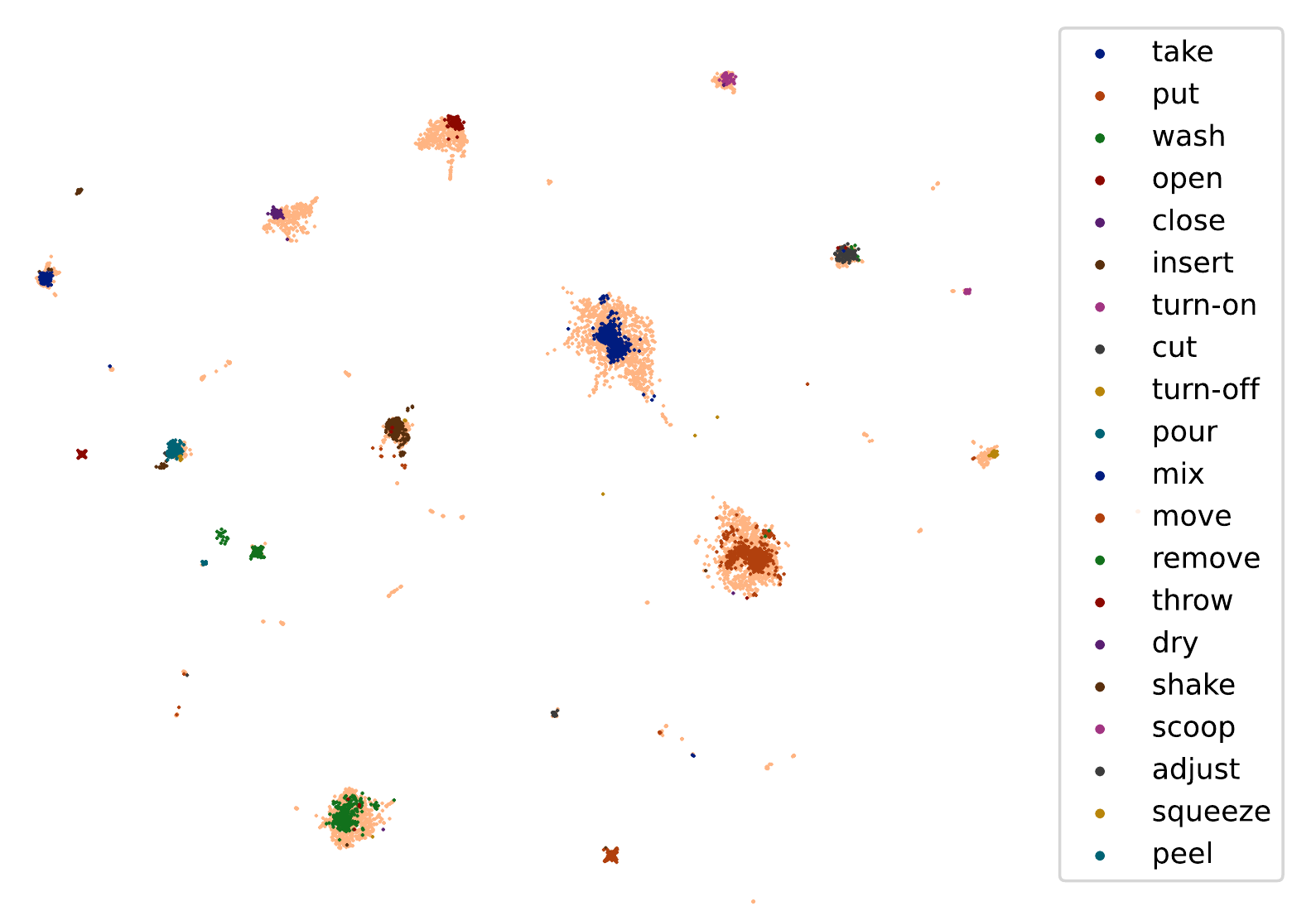}
    \subcaption{Verb embedding}
    \end{minipage}
    \vspace{1mm}
    \caption{UMAP visualisation of chosen target text queries and all target videos. Visualising text queries from 20 part-of-speech prototypes which have the most videos. Target videos are shown in ORANGE.}
    \label{fig:video-text-vis}
\end{figure}

 Fig~\ref{fig:video-text-vis} shows the UMAP visualisation of the target videos (ORANGE) and target text (other colours). Target-text were used as queries to evaluate text-video retrieval, but were not present during training.  For clarity, we only show the text embeddings from the 20 verb/noun prototypes with the most videos. Target text is aligned with target video clusters.


\section{Pseudo-label Accuracy for Alignment losses}
\label{sec:pseudo}
\begin{figure*}[h!t]
    \centering
    \textbf{Analysis of the accuracy of the pseudo-labels assigned to target videos}\\
    \begin{minipage}{0.3\linewidth}
    \includegraphics[width=\textwidth]{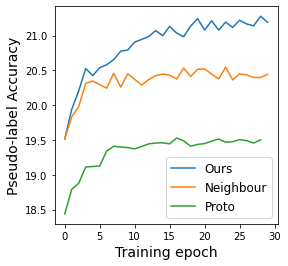}
    \subcaption{Action}
    \end{minipage}
    \begin{minipage}{0.3\linewidth}
    \includegraphics[width=\textwidth]{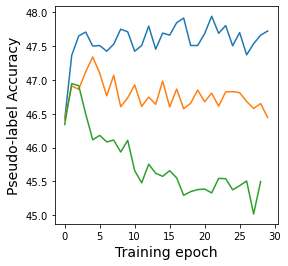}
    \subcaption{Verb}
    \end{minipage}
    \begin{minipage}{0.3\linewidth}
    \includegraphics[width=\textwidth]{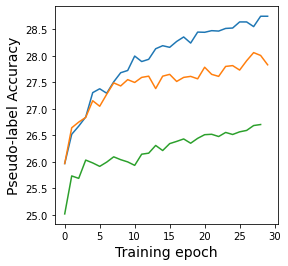}
    \subcaption{Noun}
    \end{minipage}
    \vspace{1mm}
    \caption{Pseudo-label accuracy for target videos in the action, verb and noun embedding as training progresses. Ours (BLUE) produces more accurate labels as training progresses compared to: Neighbour (ORANGE) which uses the distance to the nearest source video to determine confidence, and Proto (GREEN) which uses the nearest source prototype to propagate labels onto target videos. These different approaches are compared in Table~\ref{tab:ablation} from the main paper.}
    \label{fig:pseudolabel}
\end{figure*}

Figure~\ref{fig:pseudolabel} shows the pseudo-label accuracy over training epochs. Note that pseudo-label accuracy improves over training iterations as the domains are better aligned. Ours, with the robust prototype-based confidence measure, outperforms using the distance to the nearest source video as the confidence measure (Neighbour). Pseudo-labelling based on the nearest source prototype (Proto) incorrectly labels more target examples as it is affected by outliers.

\section{Analysis of the diversity of target video pseudo-labels}
\label{sec:variety}
 \begin{figure*}
    \centering
    \textbf{Analysis of the diversity of target video pseudo-labels}\\
    \begin{minipage}{0.3\linewidth}
    \includegraphics[width=\textwidth]{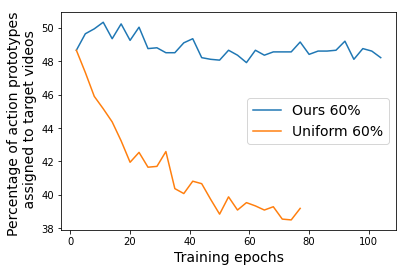}
    \subcaption{Action}
    \end{minipage}
    \begin{minipage}{0.3\linewidth}
    \includegraphics[width=\textwidth]{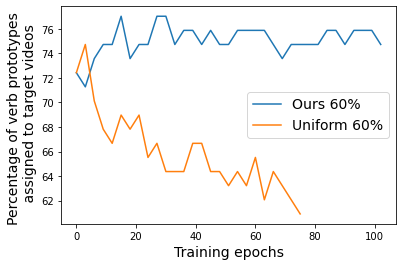}
    \subcaption{Verb}
    \end{minipage}
    \begin{minipage}{0.3\linewidth}
    \includegraphics[width=\textwidth]{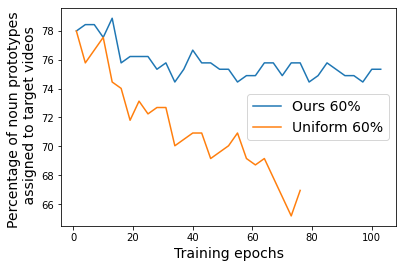}
    \subcaption{Noun}
    \end{minipage}
    \vspace{1mm}
    \caption{The percentage of prototypes (pseudo-labels) assigned to target videos in the action, verb and noun embedding as training progresses. We see that Ours (BLUE) maintains a greater variety of labels than uniform sampling (ORANGE).  
    }
    \label{fig:variety}
\end{figure*}

Figure~\ref{fig:variety} shows the percentage of prototypes  assigned to all target examples as pseudo-labels. Our proposed sampling strategy ensures that the target videos are 
assigned to a large variety of action/pos prototypes, while using uniform sampling favors assignments to prototypes corresponding to large clusters introducing an increasing  bias towards them.

\section{mAP performance}
\label{sec:map}
Tables~\ref{tab:ablation} shows the Mean Average Precision metric for the different pseudo-labelling and sampling strategies. In order to produce a binary relevancy, videos are only considered relevant to a query text if the original relevancy is strictly greater than 0.5. Similar to nDCG, Ours outperforms the alternative pseudo-labelling strategies.
\begin{table}[h!]
    \small
    \centering
    \begin{subtable}{.3\linewidth}
    \centering
    \caption{Labelling Method}
    \begin{tabular}{lr}
    \toprule
    Method & Test   \\ \midrule
    Proto & $6.21 \pm 0.02$   \\
    Ours  & $6.34 \pm 0.05$  \\
    \bottomrule
    \label{subtab:labelling}
    \end{tabular}
    \end{subtable}
    \begin{subtable}{.3\linewidth}
    \centering
    \caption{Confidence Method}
    \begin{tabular}{lr}
    \toprule
    Method & Test     \\ \midrule
    Neighbour    &  $6.09\pm 0.05$   \\
    Ours &  $6.34 \pm 0.05$   \\
    \bottomrule
    \label{subtab:confidence}
    \end{tabular}
    \end{subtable}
    \begin{subtable}{.38\linewidth}
    \centering
    \caption{Video Sampling Method}
    \begin{tabular}{lr}
    \toprule
    Method          & Test     \\ \midrule
    Uniform $(60\%)$ &  $6.28 \pm 0.02$   \\
    Ours $(60\%)$  &  $6.34 \pm 0.05$   \\
    \bottomrule
    \label{subtab:sampling}
    \end{tabular}
    \end{subtable}
    \caption{Ablation Studies on Train (mAP). Mean and standard deviation over 3 runs. We report comparable nDCG results in the main paper.
    \label{tab:ablation}}
\end{table}


\section{Additional Implementation Details}
\label{sec:furtherimplementation}

\myparagraph{Architecture and Input Features}
Video input features, $V$, are extracted from a TBN~\cite{kazakos2019TBN} model pre-trained on the source domain videos from Train and Val splits. 
The 3072 dimensional feature vector is a concatenation of RGB, Flow and Audio features---where audio is the natural sound recorded with the video. The text features are a Word2Vec~\cite{mikolov2013efficient} descriptor of length 200, trained on the Wikipedia corpus~\cite{wikicorpus}.


The base architecture is defined in JPOSE~\cite{wray2019fine} where $f(v_i)$ and $g(w_i)$ are multi-layer perceptrons with hidden layer sizes of 228 and 1664, respectively. The output video and text embeddings are vectors of length 256. The action descriptor is of length 512, which is the concatenation of the video and text embeddings.

For optimisation, weights are optimised by SGD with a learning rate of 0.01 and momentum of 0.9. We also adopt a hard-negative mining strategy for the source-domain losses. This considers only negative examples from the nearest $30\%$ of action prototypes to the action prototype of the query. This decreased the number of epochs required for training the Source-Only model, in addition to a small gain in text-video retrieval performance.

\myparagraph{Hyper-parameter Selection}
The weights for the alignment losses, $\lambda^{\times}_{s,t} = \lambda^{\times}_{t,s} = 0.1$ were found best when tuning on  Val  with values $\lambda^{\times}_{s,t}, \lambda^{\times}_{t,s} \in \{0.01,0.1,1.0\}$. The weights of the source losses $\lambda_{s}$ were those defined in JPOSE~\cite{wray2019fine}. The distance metric $d$ used for pseudo-labelling and confidence measures was the cosine distance.

As often in DA, we first pre-train our network on source and use it to initialise the proposed domain alignment network. Then we train the proposed model with cross-modal and cross-domain ranking losses, where the source-target associations are obtained with the proposed target video
pseudo-labelling and sampling strategy. At the end of each epoch we update the relevance sets. 


To obtain results on the test set (Train Target), we follow the same procedure, \textit{i.e.} we initialise our network with the model pre-trained on Train Source (note that we have different source sets for Val and Train). The network is then trained with all losses using the hyper-parameters selected from the Val set. The network is trained for $n$ epochs, where $n$ indicates the number of epochs used to train the best performing model on the Val set.
%


\myparagraph{Baselines}The shallow method, PDS, 
is applied as a pre-processing step to align $V^s$ and $V^t$. This is done only once at the beginning of the training process. In spite of its simplicity, PDS allows initial  pseudo-labelling to be improved, when compared to Source-Only. 

CORAL is also applied as a pre-processing, transforming the source input features to have the same covarience as the target.

Deep alignment methods, MMD and GRL, are optimised jointly with the source cross-modal embedding.
The weighting of the MMD loss, $\lambda_{mmd}=0.01$, and GRL, $\lambda_{GRL}=0.0001$ were found using a grid-search in the range of $[0.0001, 0.001,0.01,0.1,1.0]$. 

The conditional alignment approaches (MSTN, TPN and CDD) utilise our proposed pseudo-labelling strategy. First the networks are trained on source, similarly to Ours. For these baselines, we implement class-aware sampling~\cite{kang2019contrastive} which ensures sufficient number of examples for each prototype. This samples 30 instances from 32 different classes each mini-batch. For classes containing less than 30 instances all instances of the class are used and classes containing less than 3 samples are not sampled. The weighting for each loss is $\lambda_{MSTN}=0.1$, $\lambda_{TPN}=0.1$, $\lambda_{CDD}=0.005$ for MSTN, TPN and CDD, repsectively. These were found using a grid-search in the range of $[0.0001,0.001, 0.005, 0.01,0.1,1.0]$. 

\small
\bibliographystyle{abbrv}
\bibliography{egbib}

\begin{thebibliography}{10}

\bibitem{wikicorpus}
The wikipedia corpus.
\newblock Accessed:2020-11-11.

\bibitem{agarwal2020unsupervised}
N.~Agarwal, Y.-T. Chen, B.~Dariush, and M.-H. Yang.
\newblock Unsupervised domain adaptation for spatio-temporal action
  localization.
\newblock {\em Brit. Mach. Vis. Conf.}, 2020.

\bibitem{ak2018learning}
K.~E. Ak, A.~A. Kassim, J.~H. Lim, and J.~Y. Tham.
\newblock Learning attribute representations with localization for flexible
  fashion search.
\newblock In {\em IEEE Conf. Comput. Vis. Pattern Recog.}, 2018.

\bibitem{bak2018domain}
S.~Bak, P.~Carr, and J.-F. Lalonde.
\newblock Domain adaptation through synthesis for unsupervised person
  re-identification.
\newblock In {\em Eur. Conf. Comput. Vis.}, 2018.

\bibitem{chen2019progressive}
C.~Chen, W.~Xie, W.~Huang, Y.~Rong, X.~Ding, Y.~Huang, T.~Xu, and J.~Huang.
\newblock Progressive feature alignment for unsupervised domain adaptation.
\newblock In {\em IEEE Conf. Comput. Vis. Pattern Recog.}, 2019.

\bibitem{chen2019temporal}
M.-H. Chen, Z.~Kira, G.~AlRegib, J.~Woo, R.~Chen, and J.~Zheng.
\newblock Temporal attentive alignment for large-scale video domain adaptation.
\newblock In {\em Int. Conf. Comput. Vis.}, 2019.

\bibitem{Chen_2020_WACV}
M.-H. Chen, B.~Li, Y.~Bao, and G.~AlRegib.
\newblock Action segmentation with mixed temporal domain adaptation.
\newblock In {\em IEEE Winter Conf. on App. of Comput. Vis}, 2020.

\bibitem{chen2020action}
M.-H. Chen, B.~Li, Y.~Bao, G.~AlRegib, and Z.~Kira.
\newblock Action segmentation with joint self-supervised temporal domain
  adaptation.
\newblock In {\em IEEE Conf. Comput. Vis. Pattern Recog.}, 2020.

\bibitem{chen2020fine}
S.~Chen, Y.~Zhao, Q.~Jin, and Q.~Wu.
\newblock Fine-grained video-text retrieval with hierarchical graph reasoning.
\newblock In {\em IEEE Conf. Comput. Vis. Pattern Recog.}, 2020.

\bibitem{ChenICCV19InstanceGuidedContextRenderingDAReID}
Y.~Chen, X.~Zhu, and S.~Gong.
\newblock {Instance-Guided Context Rendering for Cross-Domain Person
  Re-Identification}.
\newblock In {\em Int. Conf. Comput. Vis.}, 2019.

\bibitem{ShuffleAttendECCV2020}
J.~Choi, G.~Sharma, S.~Schulter, and J.-B. Huang.
\newblock Shuffle and attend: Video domain adaptation.
\newblock In {\em Eur. Conf. Comput. Vis.}, 2020.

\bibitem{CsurkaTASKCV14DomainDomainSpecificClassMeans}
G.~Csurka, B.~Chidlovskii, and F.~Perronnin.
\newblock {Domain Adaptation with a Domain Specific Class Means Classifier}.
\newblock In {\em Eur. Conf. Comput. Vis. Worksh.}, 2014.

\bibitem{Damen2020RESCALING}
D.~Damen, H.~Doughty, G.~M. Farinella, A.~Furnari, J.~Ma, E.~Kazakos,
  D.~Moltisanti, J.~Munro, T.~Perrett, W.~Price, and M.~Wray.
\newblock Rescaling egocentric vision.
\newblock {\em CoRR}, 2020.

\bibitem{DamodaranECCV18DeepJDOTOptimalTransportUDA}
B.~B. Damodaran, B.~Kellenberger, R.~Flamary, D.~Tuia, and N.~Courty.
\newblock {DeepJDOT: Deep Joint Distribution Optimal Transport for Unsupervised
  Domain Adaptation}.
\newblock In {\em Eur. Conf. Comput. Vis.}, 2018.

\bibitem{deng2020rethinking}
W.~{Deng}, L.~{Zheng}, Y.~{Sun}, and J.~{Jiao}.
\newblock Rethinking triplet loss for domain adaptation.
\newblock {\em IEEE Trans. Circuit Syst. Video Technol.}, 2020.

\bibitem{DengCVPR18ImageImageDAwithPreservedSelfSimilarityReID}
W.~Deng, L.~Zheng, Q.~Ye, G.~Kang, Y.~Yang, and J.~Jiao.
\newblock {Image-Image Domain Adaptation with Preserved Self-Similarity and
  Domain-Dissimilarity for Person Re-identification}.
\newblock In {\em IEEE Conf. Comput. Vis. Pattern Recog.}, 2018.

\bibitem{DengICCV19ClusterAlignmentTeacherUDA}
Z.~Deng, Y.~Luo, and J.~Zhu.
\newblock {Cluster Alignment with a Teacher for Unsupervised Domain
  Adaptation}.
\newblock In {\em Int. Conf. Comput. Vis.}, 2019.

\bibitem{dong2021dual}
J.~Dong, X.~Li, C.~Xu, X.~Yang, G.~Yang, X.~Wang, and M.~Wang.
\newblock Dual encoding for video retrieval by text.
\newblock {\em IEEE Transactions on Pattern Analysis and Machine Intelligence},
  2021.

\bibitem{FanTOMM18UnsupervisedPersonReIDClusteringFineTuning}
H.~Fan, L.~Zheng, C.~Yan, and Y.~Yang.
\newblock {Unsupervised Person Re-identification: Clustering and Fine-tuning}.
\newblock {\em ACM Trans. Multimedia Comput. Commun. Appl.}, 2018.

\bibitem{FuICCV19SelfSimilarityGroupingDAReID}
Y.~Fu, Y.~Wei, G.~Wang, Y.~Zhou, H.~Shi, and T.~S. Huang.
\newblock {Self-similarity Grouping: A Simple Unsupervised Cross Domain
  Adaptation Approach for Person Re-identification}.
\newblock In {\em Int. Conf. Comput. Vis.}, 2019.

\bibitem{gabeur2020multi}
V.~Gabeur, C.~Sun, K.~Alahari, and C.~Schmid.
\newblock Multi-modal transformer for video retrieval.
\newblock In {\em Eur. Conf. Comput. Vis.}, 2020.

\bibitem{GaninJMLR16DomainAdversarialNN}
Y.~Ganin, E.~Ustinova, H.~Ajakan, P.~Germain, H.~Larochelle, F.~Laviolette,
  M.~Marchand, and V.~Lempitsky.
\newblock {Domain-Adversarial Training of Neural Networks}.
\newblock {\em {Advances in Computer Vision and Pattern Recognition}}, 2016.

\bibitem{ge2020mutual}
Y.~Ge, D.~Chen, and H.~Li.
\newblock Mutual mean-teaching: Pseudo label refinery for unsupervised domain
  adaptation on person re-identification.
\newblock In {\em Int. Conf. Learn. Represent.}, 2020.

\bibitem{ge2020structured}
Y.~Ge, F.~Zhu, R.~Zhao, and H.~Li.
\newblock Structured domain adaptation for unsupervised person
  re-identification.
\newblock {\em arXiv}, 2020.

\bibitem{gordo2017beyond}
A.~Gordo and D.~Larlus.
\newblock Beyond instance-level image retrieval: Leveraging captions to learn a
  global visual representation for semantic retrieval.
\newblock In {\em IEEE Conf. Comput. Vis. Pattern Recog.}, 2017.

\bibitem{hahn2019action2vec}
M.~Hahn, A.~Silva, and J.~M. Rehg.
\newblock Action2vec: A crossmodal embedding approach to action learning.
\newblock In {\em Brit. Mach. Vis. Conf.}, 2019.

\bibitem{iscen2019label}
A.~Iscen, G.~Tolias, Y.~Avrithis, and O.~Chum.
\newblock Label propagation for deep semi-supervised learning.
\newblock In {\em IEEE Conf. Comput. Vis. Pattern Recog.}, 2019.

\bibitem{jamal2018deep}
A.~Jamal, V.~P. Namboodiri, D.~Deodhare, and K.~Venkatesh.
\newblock Deep domain adaptation in action space.
\newblock In {\em Brit. Mach. Vis. Conf.}, 2018.

\bibitem{kang2019contrastive}
G.~Kang, L.~Jiang, Y.~Yang, and A.~G. Hauptmann.
\newblock Contrastive adaptation network for unsupervised domain adaptation.
\newblock In {\em Proceedings of the IEEE/CVF Conference on Computer Vision and
  Pattern Recognition}, pages 4893--4902, 2019.

\bibitem{kazakos2019TBN}
E.~Kazakos, A.~Nagrani, A.~Zisserman, and D.~Damen.
\newblock Epic-fusion: Audio-visual temporal binding for egocentric action
  recognition.
\newblock In {\em Int. Conf. Comput. Vis.}, 2019.

\bibitem{lei2021less}
J.~Lei, L.~Li, L.~Zhou, Z.~Gan, T.~L. Berg, M.~Bansal, and J.~Liu.
\newblock Less is more: Clipbert for video-and-language learning via sparse
  sampling.
\newblock In {\em IEEE Conf. Comput. Vis. Pattern Recog.}, pages 7331--7341,
  2021.

\bibitem{liu2019use}
Y.~Liu, S.~Albanie, A.~Nagrani, and A.~Zisserman.
\newblock Use what you have: Video retrieval using representations from
  collaborative experts.
\newblock {\em arXiv}, 2019.

\bibitem{Liu_2021_CVPR}
Y.~Liu, Q.~Chen, and S.~Albanie.
\newblock Adaptive cross-modal prototypes for cross-domain visual-language
  retrieval.
\newblock In {\em Proceedings of the IEEE/CVF Conference on Computer Vision and
  Pattern Recognition (CVPR)}, pages 14954--14964, June 2021.

\bibitem{LongICML15LearningTransferableFeaturesDAN}
M.~Long, Y.~Cao, J.~Wang, and M.~Jordan.
\newblock {Learning Transferable Features with Deep Adaptation Networks}.
\newblock In {\em Int. Conf. on Mach. Learn.}, 2015.

\bibitem{LongNIPS18ConditionalAdversarialDomainAdaptation}
M.~Long, Z.~Cao, J.~Wang, and M.~Jordan.
\newblock {Conditional Adversarial Domain Adaptation}.
\newblock In {\em Adv. Neural Inform. Process. Syst.}, 2018.

\bibitem{Miech_2021_CVPR}
A.~Miech, J.-B. Alayrac, I.~Laptev, J.~Sivic, and A.~Zisserman.
\newblock Thinking fast and slow: Efficient text-to-visual retrieval with
  transformers.
\newblock In {\em Proceedings of the IEEE/CVF Conference on Computer Vision and
  Pattern Recognition (CVPR)}, June.

\bibitem{miech2020end}
A.~Miech, J.-B. Alayrac, L.~Smaira, I.~Laptev, J.~Sivic, and A.~Zisserman.
\newblock End-to-end learning of visual representations from uncurated
  instructional videos.
\newblock In {\em IEEE Conf. Comput. Vis. Pattern Recog.}, 2020.

\bibitem{miech2018learning}
A.~Miech, I.~Laptev, and J.~Sivic.
\newblock Learning a text-video embedding from incomplete and heterogeneous
  data.
\newblock {\em arXiv}, 2018.

\bibitem{miech2019howto100m}
A.~Miech, D.~Zhukov, J.-B. Alayrac, M.~Tapaswi, I.~Laptev, and J.~Sivic.
\newblock Howto100m: Learning a text-video embedding by watching hundred
  million narrated video clips.
\newblock In {\em Int. Conf. Comput. Vis.}, 2019.

\bibitem{mikolov2013efficient}
T.~Mikolov, K.~Chen, G.~Corrado, and J.~Dean.
\newblock Efficient estimation of word representations in vector space.
\newblock {\em Int. Conf. Learn. Represent.}, 2013.

\bibitem{mithun2018learning}
N.~C. Mithun, J.~Li, F.~Metze, and A.~K. Roy-Chowdhury.
\newblock Learning joint embedding with multimodal cues for cross-modal
  video-text retrieval.
\newblock In {\em {Int. Conf. on Multimedia Retrieval}}, 2018.

\bibitem{Munro_2020_CVPR}
J.~Munro and D.~Damen.
\newblock Multi-modal domain adaptation for fine-grained action recognition.
\newblock In {\em IEEE Conf. Comput. Vis. Pattern Recog.}, 2020.

\bibitem{pan2019transferrable}
Y.~Pan, T.~Yao, Y.~Li, Y.~Wang, C.-W. Ngo, and T.~Mei.
\newblock Transferrable prototypical networks for unsupervised domain
  adaptation.
\newblock In {\em Proceedings of the IEEE/CVF Conference on Computer Vision and
  Pattern Recognition}, pages 2239--2247, 2019.

\bibitem{peng2019unsupervised}
Y.~Peng and J.~Chi.
\newblock Unsupervised cross-media retrieval using domain adaptation with scene
  graph.
\newblock {\em IEEE Trans. Circuit Syst. Video Technol.}, 2019.

\bibitem{planamente2021cross}
M.~Planamente, C.~Plizzari, E.~Alberti, and B.~Caputo.
\newblock Cross-domain first person audio-visual action recognition through
  relative norm alignment.
\newblock {\em ArXiv}, 2021.

\bibitem{roy2019unsupervised}
S.~Roy, A.~Siarohin, E.~Sangineto, S.~R. Bulo, N.~Sebe, and E.~Ricci.
\newblock Unsupervised domain adaptation using feature-whitening and consensus
  loss.
\newblock In {\em IEEE Conf. Comput. Vis. Pattern Recog.}, 2019.

\bibitem{pmlr-v70-saito17a}
K.~Saito, Y.~Ushiku, and T.~Harada.
\newblock Asymmetric tri-training for unsupervised domain adaptation.
\newblock In {\em Proc. Mach. Learn. Res}, 2017.

\bibitem{SenerNIPS16LearningTransferrableRepresentationsUDA}
O.~Sener, H.~Song, A.~Saxena, and S.~Savarese.
\newblock {Learning Transferrable Representations for Unsupervised Domain
  Adaptation}.
\newblock In {\em Adv. Neural Inform. Process. Syst.}, 2016.

\bibitem{Song_2021_CVPR}
X.~Song, S.~Zhao, J.~Yang, H.~Yue, P.~Xu, R.~Hu, and H.~Chai.
\newblock Spatio-temporal contrastive domain adaptation for action recognition.
\newblock In {\em Proceedings of the IEEE/CVF Conference on Computer Vision and
  Pattern Recognition (CVPR)}, June.

\bibitem{coralAAAI2016}
B.~Sun, J.~Feng, and K.~Saenko.
\newblock Return of frustratingly easy domain adaptation.
\newblock In {\em AAAI}, 2016.

\bibitem{SunTASKCV16DeepCORALCorrelationAlignment}
B.~Sun and K.~Saenko.
\newblock {Deep CORAL: Correlation Alignment for Deep Domain Adaptation}.
\newblock In {\em {TASK-CV Workshop}}, 2016.

\bibitem{TommasiICCV13FrustratinglyEasyDA}
T.~Tommasi and B.~Caputo.
\newblock {Frustratingly Easy {NBNN} Domain Adaptation}.
\newblock In {\em Int. Conf. Comput. Vis.}, 2013.

\bibitem{TzengCVPR17AdversarialDiscriminativeADDA}
E.~Tzeng, J.~Hoffman, K.~Saenko, and T.~Darrell.
\newblock {Adversarial Discriminative Domain Adaptation}.
\newblock In {\em IEEE Conf. Comput. Vis. Pattern Recog.}, 2017.

\bibitem{wang2016structure}
L.~{Wang}, Y.~{Li}, and S.~{Lazebnik}.
\newblock Learning deep structure-preserving image-text embeddings.
\newblock In {\em IEEE Conf. Comput. Vis. Pattern Recog.}, 2016.

\bibitem{wang2021t2vlad}
X.~Wang, L.~Zhu, and Y.~Yang.
\newblock T2vlad: global-local sequence alignment for text-video retrieval.
\newblock In {\em IEEE Conf. Comput. Vis. Pattern Recog.}, pages 5079--5088,
  2021.

\bibitem{Wray_2021_CVPR}
M.~Wray, H.~Doughty, and D.~Damen.
\newblock On semantic similarity in video retrieval.
\newblock In {\em Proceedings of the IEEE/CVF Conference on Computer Vision and
  Pattern Recognition (CVPR)}, 2021.

\bibitem{wray2019fine}
M.~Wray, D.~Larlus, G.~Csurka, and D.~Damen.
\newblock Fine-grained action retrieval through multiple parts-of-speech
  embeddings.
\newblock In {\em Int. Conf. Comput. Vis.}, 2019.

\bibitem{WuCVPR18ExploitTheUnknownGradually}
Y.~Wu, Y.~Lin, X.~Dong, Y.~Yan, W.~Ouyang, and Y.~Yang.
\newblock {Exploit the Unknown Gradually: One-Shot Video-Based Person
  Re-Identification by Stepwise Learning}.
\newblock In {\em IEEE Conf. Comput. Vis. Pattern Recog.}, 2018.

\bibitem{xie2018learning}
S.~Xie, Z.~Zheng, L.~Chen, and C.~Chen.
\newblock Learning semantic representations for unsupervised domain adaptation.
\newblock In {\em International conference on machine learning}, pages
  5423--5432. PMLR, 2018.

\bibitem{ZhangCVPR18CollaborativeAdversarialUDA}
W.~Zhang, W.~Ouyang, W.~Li, and D.~Xu.
\newblock {Collaborative and Adversarial Network for Unsupervised Domain
  Adaptation}.
\newblock In {\em IEEE Conf. Comput. Vis. Pattern Recog.}, 2018.

\bibitem{ZhangICCV19SelfTrainingWithProgressiveAugmentationDAReID}
X.~Zhang, J.~Cao, C.~Shen, and M.~You.
\newblock {Self-Training with Progressive Augmentation for Unsupervised
  Cross-Domain Person Re-Identification}.
\newblock In {\em Int. Conf. Comput. Vis.}, 2019.

\bibitem{ZhangICML19BridgingTheoryAlgorithmDA}
Y.~Zhang, T.~Liu, M.~Long, and M.~Jordan.
\newblock {Bridging Theory and Algorithm for Domain Adaptation}.
\newblock In {\em Int. Conf. on Mach. Learn.}, 2019.

\bibitem{zhang2015zero}
Z.~Zhang and V.~Saligrama.
\newblock Zero-shot learning via semantic similarity embedding.
\newblock In {\em Int. Conf. Comput. Vis.}, 2015.

\bibitem{zhou2019ladder}
M.~Zhou, Z.~Niu, L.~Wang, Z.~Gao, Q.~Zhang, and G.~Hua.
\newblock Ladder loss for coherent visual-semantic embedding.
\newblock In {\em AAAI}, 2020.

\end{thebibliography}

\end{document}